\documentclass[11pt]{article}
\usepackage{microtype}  
\usepackage{fullpage}
\usepackage{url}    
\usepackage{booktabs}       
\usepackage{nicefrac}       
\PassOptionsToPackage{dvipsnames}{xcolor}
\usepackage{xcolor,colortbl} 
\usepackage{multirow}
\usepackage{tablefootnote}
\usepackage{subcaption}
\usepackage{makecell}
\usepackage{silence}
\usepackage{preamble}
\usepackage{mathnotations}

\usepackage{widebar}

\usepackage{arxiv}

\numberwithin{equation}{section}

\begin{document}
\title{\huge Differentially Private Policy Optimization in Linear MDPs with Adversarial Rewards}
\author{Yi He \\[0.4em]\normalsize Wayne State University, Detroit, USA}
\date{}
\maketitle

\begin{abstract}
Reinforcement learning (RL) in personalized services often relies on sensitive interaction data. Motivated by the need to protect these data, we study differentially private (DP) policy optimization in episodic linear Markov decision processes (MDPs) with unknown transitions and adaptive adversarial linear rewards under full-information feedback. This setting poses two key challenges: past trajectories are repeatedly reused for policy evaluation, and reward choices may depend on previously released private statistics. The proposed algorithm combines batched proximal policy optimization (PPO) updates with optimistic policy evaluation, using private regression statistics and reward estimates to protect both trajectories and reward functions under joint differential privacy (JDP). With all other parameters fixed, our high-probability regret bound retains the same leading $\widetilde{\mathcal{O}}\left(K^{3/4}\right)$ dependence on the number of episodes $K$ as the corresponding non-private bound for batched optimistic PPO. The additional privacy terms scale as $\widetilde{\mathcal{O}} \left(K^{5/8}\right)$ and $\widetilde{\mathcal{O}} \left(K^{1/2}\right)$. Thus, privacy incurs only lower-order additive costs in $K$, even under adaptive adversarial rewards.
\end{abstract}

\section{Introduction}

Reinforcement learning (RL) provides a principled framework for sequential decision making in unknown environments. It is widely used in personalized systems, with applications ranging from online recommendations~\citep{li2010contextual} and personalized healthcare~\citep{yu2021reinforcement} to large language models (LLMs) fine-tuned via reinforcement learning from human feedback (RLHF)~\citep{ouyang2022training}. Among RL methods, policy optimization (PO), such as proximal policy optimization (PPO)~\citep{schulman2017proximal} and group relative policy optimization (GRPO)~\citep{shao2024deepseekmath}, is one of the most widely used approaches in practice. We focus on a setting in which the learner interacts with a sequence of users. Each user generates one episode of interaction and evaluates its outcome according to a personal reward function.

While highly effective, RL-based personalization can raise privacy concerns, since the data collected in each episode often reflects sensitive user information. For example, in personalized medical care, the states may encode a patient's medical history, while the reward reflect treatment effectiveness or adverse reactions. In recommendation and LLM applications, the reward function may reflect a user's preference, which can reveal personal interests, intentions, or beliefs. Indeed, existing non-private RL methods have been shown to expose sensitive information about training data or the environments~\citep{pan2019you, feng2025exposingprivacygapsmembership,liu2025grpoprivacyriskmembership}. Thus, it is essential to protect not only the trajectories of users, but also their reward functions.

Reward functions may vary substantially across episodes because users have different preferences. Moreover, the outputs released by the learner may in turn influence the preferences of later users. Adversarial MDPs~\citep{even2009online, neu2010online, zimin2013online} capture both aspects by allowing the reward functions to change arbitrarily across episodes and, under an adaptive adversary, to depend on previously released outputs. PO is well suited to this setting, since its mirror-descent updates naturally handle changing rewards. In particular, optimistic PPO achieves sublinear regret with adversarial rewards under linear function approximation~\citep{cai2020provablyefficientexplorationpolicy, zhong2023theoreticalanalysisoptimisticproximal}.

Motivated by privacy concerns, a recent line of theoretical work has studied differential privacy (DP)~\citep{dwork2014algorithmic} in online RL. Since requiring the actions recommended to a user to be insensitive to that user's own data can force linear regret in general RL settings, this literature mainly adopts joint differential privacy (JDP)~\citep{kearns2014mechanism, vietri2020private}. Sublinear regret under JDP has been established in tabular MDPs~\citep{vietri2020private, chowdhury2021differentiallyprivateregretminimization, qiao2023nearoptimaldifferentiallyprivatereinforcement}, with linear function approximation~\citep{zhou2022differentially, luyo2021differentiallyprivateexplorationreinforcement}, and with general function approximation~\citep{he2026towards}. For PO,~\citet{he2025samplecomplexitydifferentiallyprivate} show that the privacy cost often appears as a lower-order term in the sample complexity. These results assume stochastic rewards. For adversarial rewards,~\citet{bai2024differentially} study tabular MDPs against an oblivious adversary and obtain expected regret bounds. Their guarantees do not cover large state spaces, where function approximation is necessary, or adversaries that react to the outputs released by the learner.

Extending private RL to adaptive adversarial rewards raises two challenges. First, batched optimistic PPO repeatedly reuses past trajectories for policy evaluation. At each batch update, the aggregate regression target vectors are recomputed from past transitions using the current value function and released with fresh noise. Each trajectory therefore enters many releases, and the privacy loss accumulates under composition. Second, the adversary may choose the reward functions based on previously released private statistics. The rewards then depend on the private data, so the privacy analysis must account for reward choices that change with the released outputs. Since the rewards are no longer fixed in advance, the regret analysis must also hold uniformly over all reward sequences that the adversary may select. These challenges lead to the following question:
\begin{center}
    \emph{Can private policy optimization with adaptive adversarial rewards retain the leading-order regret of non-private batched optimistic PPO}?
\end{center}

We answer this question in the affirmative for linear MDPs. Our main contributions are as follows.

\begin{enumerate}[leftmargin=*, itemsep=0.5em]
    \item We study DP policy optimization in episodic linear MDPs with unknown transitions and adaptive adversarial linear rewards under full-information feedback. Both the trajectories and the reward functions are protected under JDP. We propose a private PO algorithm (Algorithm~\ref{alg:privPO}) with a high-probability regret bound of order $\widetilde{\cO}(K^{3/4})$ (Theorem~\ref{thm:reg}). With all other parameters fixed, the leading dependence on the number of episodes $K$ matches the non-private bound of batched optimistic PPO~\citep{zhong2023theoreticalanalysisoptimisticproximal}, and privacy contributes only lower-order terms. To the best of our knowledge, this is the first regret guarantee for differentially private RL in linear MDPs with adversarial rewards.

    \item Our algorithm builds on batched optimistic PPO~\citep{zhong2023theoreticalanalysisoptimisticproximal} and uses private regression statistics and reward estimates. Batching controls the privacy cost of reusing past trajectories, while reward averaging reduces the noise in reward releases. At a fixed public history, only the replaced user's coefficients can change a batch average, even though subsequent rewards may differ across runs. For regret, we extend the covering argument by fixing the independent reward noise and ranging over all admissible batch-average coefficients, including those selected adaptively.
\end{enumerate}

\subsection{Related Work}

We mainly discuss the most relevant work here and give a detailed discussion in Appendix~\ref{sec:related}. For linear MDPs with adversarial rewards and full-information feedback,~\citet{zhong2023theoreticalanalysisoptimisticproximal} establish $\widetilde{\cO}(K^{3/4})$ regret without requiring linear rewards. When rewards are linear in the transition feature,~\citet{sherman2024rateoptimalpolicyoptimizationlinear} and~\citet{cassel2024warm} obtain $\widetilde{\cO}(\sqrt{K})$ regret against oblivious and adaptive adversaries, respectively. Our bound remains $\widetilde{\cO}(K^{3/4})$ in this shared-feature setting. Cassel and Rosenberg recompute regression targets from past transitions in every episode, so a direct private implementation must account for repeated releases of statistics computed from the same data. We build on~\citet{zhong2023theoreticalanalysisoptimisticproximal}'s batched optimistic PPO to limit these releases.

Under fixed expected rewards,~\citet{zhou2022differentially} establish JDP regret guarantees for value iteration and policy optimization in linear mixture MDPs, while~\citet{luyo2021differentiallyprivateexplorationreinforcement} study private exploration in linear and linear mixture MDPs. \citet{he2025samplecomplexitydifferentiallyprivate} analyze the sample complexity of private policy optimization. For adversarial rewards, the closest work is~\citet{bai2024differentially}, which gives expected regret guarantees for tabular MDPs against an oblivious adversary. Our work considers linear function approximation and adaptive adversaries, with high-probability regret guarantees.

\section{Preliminaries}

\subsection{Linear MDPs with adversarial linear rewards}

We consider an episodic MDP with adversarial rewards and full-information feedback. The state space $\cS$ and action space $\cA$ are finite, with cardinalities $S := \abs{\cS}$ and $A:=\abs{\cA}$. We assume $A \ge 2$, since the regret is identically zero when $A=1$. The learner interacts with the environment for $K$ episodes, each consisting of $H$ stages, where $H$ and $K$ are positive integers. The unknown transition kernels $\cP:=\{ \cP_h \}_{h=1}^H$ may depend on the stage $h$ but remain fixed across episodes. $\cP_h(\cdot \mid x,a)$ is the distribution of the next state given state $x$ and action $a$ at stage $h$.


\begin{definition}[Adversarial linear rewards] \label{def:adversarial-linear}
     Let $p \in \mathbb{N}$ and $L >0$. Let the known reward feature $\psi : \cS \times \cA \to \mathbb{R}^{p}$ satisfy $\norm{\psi (x,a)}_2 \le 1$. Before episode $k$ begins, the adversary selects $\theta_h^{(k)} \in \mathbb{R}^p$ for every $h \in [H]$, with $\norm{\theta_h^{(k)}}_2 \le L$, and for all $(x,a) \in \cS \times \cA$,
     \begin{align*}
         r_h^{(k)} (x,a) = \inner{\psi(x,a)}{\theta_h^{(k)}} \in [0,1].
     \end{align*}
\end{definition}


The linear reward assumption above is motivated by the cost of protecting entire reward functions. A direct Gaussian release of arbitrary reward tables must be calibrated to an $\ell_2$ sensitivity proportional to $\sqrt{SA}$. This factor enters the regret bound and makes the guarantee unsuitable for large state and action spaces. With a known linear reward representation and bounded coefficients, we instead privatize the batch-average coefficients, leaving only logarithmic dependence on $S$ and $A$ in the reward-noise term.

\begin{remark}
    The adversary may choose the reward coefficients based on previously released privatized statistics and its own independent randomness. It cannot condition this choice on the current episode's trajectory or unreleased private observations. Thus, the rewards may adapt across episodes, but all coefficients for episode $k$ must be fixed before that episode begins. Given the same public transcript and random seed, the adversary chooses the same rewards for a later user even if an earlier user received different actions in private.
\end{remark}

\textbf{Full-information feedback.} After each episode, the learner observes the reward functions on all state-action pairs, including pairs not visited in that episode. The next-state observations remain restricted to the realized trajectory, so the unknown transitions still have to be estimated from interaction data. The reward functions are supplied to the learner as private input; their influence on the statistics and actions released to other users is protected by JDP. With bandit feedback, the reward at a visited pair generally leaves its coefficient vector unidentified. A private algorithm would also have to control the estimation error and sensitivity of rewards inferred from the visited features.

\textbf{Policies and interaction.} A policy $\pi = (\pi_1, \ldots, \pi_H)$ is a collection of randomized Markov decision rules $\pi_h : \cS \rightarrow \Delta(\cA)$. Let $\Pi$ denote the class of all such policies. At the beginning of episode $k$, the learner chooses $\pi^{(k)} \in \Pi$ using information from previous episodes and its internal randomness. Every episode starts from the same fixed initial state $x_1^{(k)} = x_1 \in \cS$. At each stage $h \in [H]$, the learner observes $x_h^{(k)}$ and interacts according to 
\begin{align*}
    a_h^{(k)} \sim \pi_h^{(k)} \left( \cdot \mid x_h^{(k)} \right), \quad 
    x_{h+1}^{(k)} \sim \cP_h \left( \cdot \mid x_h^{(k)}, a_h^{(k)} \right).
\end{align*}

\textbf{Value functions.} For $\pi \in \Pi$, we define the value under the reward functions of episode $k$ as
\begin{align*}
    V_h^{(\pi, k)} (x) := \mathbb{E}_{\pi} \left[ \sum_{j=h}^{H} r_j^{(k)} (x_j, a_j) \mid x_h = x \right],
\end{align*}
and $V_{H+1}^{(\pi, k)} \equiv 0 $. For any $V: \cS \rightarrow \mathbb{R}$, define the transition operator by 
\begin{align*}
    (\mathbb{P}_h V)(x,a) := \sum_{x' \in \cS} \cP_h(x' \mid x,a) V(x').
\end{align*}

The action-value function $Q_h^{(\pi, k)}(x,a)$ is the expected remaining reward when the process starts at state $x$ and stage $h$, takes action $a$, and then follows $\pi$. It satisfies the Bellman equations with $\inner{\cdot}{\cdot}_{\cA}$ denoting the inner product over actions:
\begin{align*}
    Q_h^{(\pi,k)} (x,a) &= r_h^{(k)} (x,a) + \left( \mathbb{P}_h V_{h+1}^{(\pi,k)} \right)(x,a), \\
    V_h^{(\pi, k)} (x) &= \inner{Q_h^{(\pi,k)} (x,\cdot)}{\pi_h(\cdot \mid x)}_{\cA}.
\end{align*}

Since rewards lie in $[0,1]$, both $V_h^{(\pi, k)}$ and $Q_h^{(\pi,k)}$ take values in $[0, H-h+1]$.

For adaptive rewards, $V_h^{(\pi,k)}$ evaluates policy $\pi$ against the reward function realized in episode $k$, holding that function fixed. The regret below is external regret for the realized sequence, rather than a counterfactual response of the adversary to $\pi$.

\textbf{Learning objective.} The performance of the learner is measured via the regret
\begin{align*}
    \reg(K) := \max_{\pi \in \Pi} \sum_{k=1}^K \left( V_1^{(\pi,k)}(x_1) - V_1^{(\pi^{(k)}, k)} (x_1) \right).
\end{align*}

Let $\pi^{\star}$ denote a best fixed policy in hindsight,
\begin{align*}
    \pi^{\star} \in \argmax_{\pi \in \Pi}  \sum_{k=1}^K V_1^{(\pi, k)} (x_1).
\end{align*}

\textbf{Linear transitions.} We adopt the linear transition model of~\citet{jin2020provably}. 

\begin{definition}[Linear MDPs] \label{def:linearMDP}
    We say an MDP $\left(\cS, \cA, H, K, \{ r_h^{(k)} \}, \cP \right)$ is a linear MDP if there exists a known feature $\phi: \cS \times \cA \rightarrow \mathbb{R}^d$ satisfying $\norm{\phi(x,a)}_2 \le 1$ for any $(x,a) \in \cS \times \cA$, such that for any $(x,a,h) \in \cS \times \cA \times [H]$, we have 
    \begin{align*}
        \cP_h ( \cdot |x,a) = \phi(x,a)^\top \mu_h(\cdot),
    \end{align*}
    where $\mu_h = \left( \mu_h^{(1)}, \ldots, \mu_h^{(d)} \right)$ is a vector of $d$ unknown signed measures over $\cS$. Write $\inner{\mu_h}{v}_{\cS}:=\sum_{x\in\cS}\mu_h(\{x\})v(x)$. We assume that
    \begin{align*}
        \sup_{v:\cS\to[0,1]}\norm{\inner{\mu_h}{v}_{\cS}}_2\le\sqrt{d},\quad h\in[H].
    \end{align*}

\end{definition}

The reward feature $\psi$ need not coincide with the transition feature $\phi$. This allows reward preferences and transition dynamics to have different representations. The commonly used shared-feature setting is recovered by taking $\psi=\phi$ and $p=d$.

\subsection{Differential privacy} \label{sec:differential-privacy}

Following the episodic formulation in~\citet{vietri2020private}, we associate each episode $k$ with a single private record $u_k$. We use the term "user" to refer to such an episode record. In our setting, $u_k$ contains the user's state responses to all possible action histories within the episode, together with the rule selecting her full reward functions from the preceding public history. We write 
\begin{align*}
    U_K = (u_1, \ldots, u_K) \in \cU^{K}
\end{align*}
for the collection of private episode records, where $\cU$ is the space of possible records.

Fix an episode $i$. We say two inputs $U_K$ and $U_K'$ are $i$-th neighbors if $u_i$ and $u_i'$ are two admissible episode records with $u_j = u_j'$ for every $j \ne i$. The reward-selection rules of episodes $j\ne i$,  the public features $\phi, \psi$, the transition kernel, and the initial state $x_1$ are fixed in the neighbor relation.


Replacing $u_i$ may change both its state responses and its reward-selection rule, and hence the reward coefficient vector selected for episode $i$. In particular, the same public history induces the same reward choice in the two neighboring worlds for every episode $j\ne i$, whereas the reward choice for episode $i$ may differ. Once the public histories become different, later reward choices and realized trajectories may also become different as a consequence of the interaction. These differences do not
change the neighboring relation defined above.

Let $\mathsf{Tr}(U_K)$ denote the public transcript generated during the interaction, consisting of all information publicly released by the learner. Let $\cM$ denote the randomized learning mechanism. The complete output $\cM(U_K)$ consists of this transcript, the actions in all episodes, and the final policy $\widehat{\pi}$. The actions in each episode are sent to that episode's user, and the final policy is publicly released after the interaction. All privacy guarantees below are required to hold for every admissible collection of episode records, including their reward-selection rules. They hold for each fixed realization of the adversary's independent random seed.

\begin{definition}[Differential Privacy (DP)]
For $\epsilon\ge 0$ and $\delta\in[0,1]$, a randomized mechanism $\cF$ with input $U_K$ is $(\epsilon,\delta)$-differentially private if, for every episode $i$, every pair of $i$-th neighboring inputs $U_K,U'_K$, and every measurable event $\cE$ in the output space of $\cF$,
\begin{align*}
    \mathbb{P}\left(\cF(U_K)\in\cE\right)\le e^\epsilon\mathbb{P}\left(\cF(U'_K)\in\cE\right)+\delta.
\end{align*}
\end{definition}

This is the standard DP inequality~\citep{dwork2014algorithmic} under the neighboring relation defined above. In our privacy analysis, we first establish DP for the public transcript $\mathsf{Tr}$.

However, requiring DP for the complete output $\cM$ would also require the actions sent to user $i$ to satisfy the DP inequality when $u_i$ changes, while these actions can directly depend on $u_i$. We therefore adopt joint differential privacy (JDP)~\citep{kearns2014mechanism}. Let $\cM_{-i}(U_K)$ denote all outputs of the learning process observable to the adversary and users other than $i$, excluding only the actions in episode $i$. In particular, $\cM_{-i}(U_K)$ retains the public transcript $\mathsf{Tr}(U_K)$, the final policy $\widehat{\pi}$, and all actions in the other episodes.

\begin{definition}[Joint Differential Privacy (JDP)]
    For $\epsilon\ge 0$ and $\delta\in[0,1]$, the learning mechanism is $(\epsilon,\delta)$-jointly differentially private if, for every realization of the adversary's independent random seed, every episode $i$, every pair of $i$-th neighboring inputs $U_K,U'_K$, and every measurable event $\cE$ in the output space of $\cM_{-i}$, we have
    \begin{align*}
        \mathbb{P}\left(\cM_{-i}(U_K)\in\cE\right)\le e^\epsilon\mathbb{P}\left(\cM_{-i}(U'_K)\in\cE\right)+\delta.
    \end{align*}
\end{definition}

We prove the JDP guarantee by first establishing DP for the public transcript $\mathsf{Tr}$. The same proof continues to hold after fixing the learner's independent action-sampling random seeds. Since these seeds are independent of the private episode records, and the DP bound holds for every fixed choice of the seeds, releasing the transcript together with the seeds also satisfies DP. Conditional on the transcript and these random seeds, the actions shown to every user $j \ne i$ depend only on her unchanged episode record $u_j$. The JDP guarantee then follows from the billboard argument~\citep{hsu2016}.

\section{Private Policy Optimization with Adversarial Rewards}

In this section, we present a private policy optimization algorithm for linear MDPs with adversarial linear rewards. We divide the $K$ episodes into $M = K/B$ batches\footnote{For simplicity, we assume that $B$ is a positive integer that divides $K$, so that $M=K/B$ is an integer.}, and let $t_m = (m-1)B + 1$ denote the first episode of batch $m$. We set $t_{M+1}:=K+1$, so that batch $m$ consists of episodes $t_m,\ldots,t_{m+1}-1$. At the beginning of each batch, the algorithm updates the policy using the previous private $Q$-function estimates and then evaluates the updated policy using private statistics. The policy remains fixed within the batch. 

The adversary can still choose a different reward function in every episode within a batch. Batching restricts only how often the learner updates its policy and recomputes regression targets. At a batch boundary, all rewards from the preceding batch are available, and their average is used for the next policy evaluation. This delay trades immediate policy adaptation for fewer regression releases and a smaller sensitivity of reward averages.

\begin{algorithm}[!ht]
  \caption{Private Policy Optimization under Adversarial MDPs}
  \label{alg:privPO}
\small
\begin{algorithmic}[1]
\raggedright
    \STATE {\bfseries Input:} batch size $B \in \mathbb{N}$, learning rate $\alpha > 0$, regularization parameter $\lambda > 0$, confidence radius $\beta_{\mathsf{priv}}>0 $, privacy parameters $\epsilon, \delta$, and noise scales $\sigma_{\mathsf{node}},\sigma_u,\sigma_{\theta}>0$ chosen accordingly. Here $E_h^{(k)}$ is the noise from the tree nodes covering $[1,k-1]$, and the fresh noises have laws $\zeta_{u,h}^{(k)}\sim\cN(0,\sigma_u^2 I_d)$ and $\zeta_{\theta,h}^{(k)}\sim\cN(0,\sigma_\theta^2 I_p)$.
    
    \STATE Initialize $\widetilde{Q}_h^{(0)}$ as zero functions, $\bar{\theta}_h^{(1)} = 0 \in \mathbb{R}^p$, $\pi_h^{(0)}$ as uniform on $\cA$, and trees $\cT_h$ \mbox{for all $h\in[H]$.}
    \STATE Set $M\!:=\!K/B$ and $t_m\!:=\!\left(m-1\right)B+1$ \mbox{for $m\in[M]$.}
    
    \FOR{$k = 1, \ldots, K$}
        \STATE Set $\widetilde{V}_{H+1}^{(k)} (\cdot) = 0$.
        \IF{ $k=t_m$ for some $m\in[M]$ }
            \STATE Update $\pi_h^{(k)}\!\propto\!\pi_h^{(k-1)}\!\cdot\!\exp\!\left\{\alpha\widetilde{Q}_h^{(k-1)}\right\}$ \mbox{for all $h\in[H]$.}
            
            \FOR{ step $h = H, H-1, \cdots, 1$}
                \STATE $\Lambda_h^{(k)}\!\leftarrow\!\sum_{\tau=1}^{k-1}\!\phi\!\left(x_h^{(\tau)},\!a_h^{(\tau)}\right)\!\phi\!\left(x_h^{(\tau)},\!a_h^{(\tau)}\right)^\top$
                \STATE $\widetilde{\Lambda}_h^{(k)}\!\leftarrow\!\Lambda_h^{(k)}\!+\!\lambda I_d\!+\!E_h^{(k)}$
                \STATE $\widetilde{u}_h^{(k)}\!\leftarrow\!\sum_{\tau=1}^{k-1}\!\phi\!\left(x_h^{(\tau)},\!a_h^{(\tau)}\right)\widetilde{V}_{h+1}^{(k)}\!\left(x_{h+1}^{(\tau)}\right)\!+\!\zeta_{u,h}^{(k)}$
                
                \IF{$\widetilde{\Lambda}_h^{(k)}\succ0$}
                    \STATE $\widetilde{w}_h^{(k)} \!\leftarrow\! \left( \widetilde{\Lambda}_h^{(k)} \right)^{-1} \widetilde{u}_h^{(k)}$
                    \STATE $\widetilde{\Gamma}_h^{(k)} \!\leftarrow\! \beta_{\mathsf{priv}} \!\cdot\! \sqrt{\phi^\top ( \widetilde{\Lambda}_h^{(k)})^{-1} \phi }$ 
                \ELSE
                    \STATE $\widetilde{w}_h^{(k)} \!\leftarrow\! 0,\quad \widetilde{\Gamma}_h^{(k)} \!\leftarrow\! 0$
                \ENDIF    
                \STATE $\widehat{\mathbb{P}}_h^{(k)} \widetilde{V}_{h+1}^{(k)} \!\leftarrow\! \min \{ \phi^\top \widetilde{w}_h^{(k)}\!+\!\widetilde{\Gamma}_h^{(k)}, H-h \}^{+}$
                \STATE If $k>1$, set $\bar{\theta}_h^{(k)} \!\leftarrow\! \frac{1}{B}\!\sum_{j=k-B}^{k-1}\!\theta_h^{(j)}$
                \STATE $\widetilde{\theta}_h^{(k)} \!\leftarrow\! \bar{\theta}_h^{(k)}\!+\!\zeta_{\theta,h}^{(k)}$
                \STATE $\widetilde{r}_h^{(k)}(x,a)\!\leftarrow\!\mathsf{Clip}_{[0,1]} \left(\inner{\psi(x,a)}{\widetilde{\theta}_h^{(k)}}\right)$
                \STATE $\widetilde{Q}_h^{(k)} \!\leftarrow\! \widetilde{r}_h^{(k)}\!+\!\widehat{\mathbb{P}}_h^{(k)} \widetilde{V}_{h+1}^{(k)}$ 
                \STATE $\widetilde{V}_h^{(k)} \!\leftarrow\! \inner{\widetilde{Q}_h^{(k)}}{\pi_h^{(k)}}_\cA$
            \ENDFOR
        \ELSE
            \STATE Reuse the policy and the reward, $Q$-function, and $V$-function estimates from episode $k-1$.
        \ENDIF

        \STATE Execute $\pi^{(k)}$ for one episode and observe \mbox{$\{r_h^{(k)}\}_{h=1}^{H}$ after it.}
        \STATE Insert $\phi\!\left(x_h^{(k)},\!a_h^{(k)}\right)\!\phi\!\left(x_h^{(k)},\!a_h^{(k)}\right)^\top$ \mbox{into $\cT_h$ for all $h\in[H]$.}
    \ENDFOR
    \STATE Return $\widehat{\pi}:=\pi^{(K)}$.
\end{algorithmic}
\end{algorithm}

\textbf{Policy improvement.} Following the proximal policy updates in~\citet{cai2020provablyefficientexplorationpolicy} and~\citet{zhong2023theoreticalanalysisoptimisticproximal}, we first describe the proximal policy update for a given $Q$-function $Q_h:\cS\times\cA\to\mathbb{R}$. For each fixed state $x$, we suppress the state argument in the policy distributions and the Q-function throughout this update. At each batch start $k = t_m$ for each state $x \in \cS$ and stage $h \in [H]$ with learning rate $\alpha > 0$, we solve:
\begin{align*}
    \pi_h^{(k)} := \argmax_{\pi_h\in\Delta(\cA)} \left\{ \inner{Q_h}{\pi_h}_{\cA} - \frac{1}{\alpha} \mathsf{KL}\left( \pi_h \| \pi_h^{(k-1)} \right) \right\}.
\end{align*}
Here $\pi_h$ is the candidate action distribution, and the KL divergence is 
\begin{align*}
    \mathsf{KL}\left( \pi_h \| \pi_h^{(k-1)} \right) := \sum_{a \in \cA} \pi_h(a) \log \frac{\pi_h(a)}{\pi_h^{(k-1)}(a)}.
\end{align*}

The first term in the objective is the expected $Q$-value under the candidate policy, while the KL penalty discourages large deviations from the previous policy. This problem has the closed-form solution
\begin{align*}
    \pi_h^{(k)}(a) = \frac{\pi_h^{(k-1)}(a) \exp\left\{\alpha Q_h(a)\right\}}{\sum_{b\in\cA}\pi_h^{(k-1)}(b) \exp\left\{\alpha Q_h(b)\right\}}.
\end{align*}

In Algorithm~\ref{alg:privPO}, we use the previous private Q-function estimate $\widetilde{Q}_h^{(k-1)}$ in place of $Q_h$.

\textbf{Policy evaluation.} To evaluate the updated policy, we compute the private estimates $\widetilde{Q}_h^{(k)}$ and $\widetilde{V}_h^{(k)}$ backward from step $H$ to 1. The expected next-step value $\mathbb{P}_h \widetilde{V}_{h+1}^{(k)}$ is linear in $\phi$, so we estimate it by regularized least squares on past transitions. We add noise to the regression statistics to protect privacy and include an exploration bonus to account for estimation error. After clipping this value estimate to $[0,H-h]$, we add the reward estimate from the preceding batch's average reward coefficients to obtain $\widetilde{Q}_h^{(k)}$. We describe the mechanisms for privatizing the covariance matrices, regression target vectors, and batch-average reward coefficients as follows. 

\textbf{Private covariance matrices.} At a batch start $k = t_m$, the algorithm forms a  covariance matrix using all previous trajectories. The same episode therefore contributes to the matrices used in several batch updates. To limit the noise needed for these repeated updates, we use the binary tree mechanism~\citep{chan2011private}, which reuses noisy sums of contributions from earlier episodes. For each stage $h$, we maintain a binary tree $\cT_h$, where episode $\tau$ contributes
\begin{align*}
    \phi(x_h^{(\tau)},a_h^{(\tau)})\phi(x_h^{(\tau)},a_h^{(\tau)})^\top.
\end{align*}

We use the dyadic intervals contained in $[1,K]$, so each episode belongs to at most one interval at each of $1+\lfloor\log_2 K\rfloor$ levels. Each node stores the sum over a contiguous interval of episodes, with a symmetric Gaussian matrix $e_{h,v}$ added to it. The upper-triangular noise entries are sampled independently from $\cN \left(0,\sigma_{\mathsf{node}}^2\right)$ and copied to the lower triangle. At a batch start $k=t_m$, we combine the noisy sums from nodes whose intervals cover episodes $1,\ldots,k-1$. This requires at most $1+\log_2K$ nodes with disjoint intervals, and each episode contributes to at most $1+\log_2K$ nodes per stage. Reusing an already noisy node does not consume additional privacy budget. Let $E_h^{(k)}$ be the sum of the selected node noises. The private covariance matrix is
\begin{align*}
    \widetilde{\Lambda}_h^{(k)} := \lambda I_d + \sum_{\tau=1}^{k-1} \phi(x_h^{(\tau)},a_h^{(\tau)})\phi(x_h^{(\tau)},a_h^{(\tau)})^\top + E_h^{(k)}.
\end{align*}

\textbf{Private target vectors.} To form the regression target, we multiply each past feature vector by the current estimated value $\widetilde{V}_{h+1}^{(k)}$ of its observed next state. At each batch start and each stage, we sum these terms and add Gaussian noise:
\begin{align*}
    \widetilde{u}_h^{(k)} &:= \sum_{\tau=1}^{k-1} \phi \left( x_h^{(\tau)}, a_h^{(\tau)} \right) \widetilde{V}_{h+1}^{(k)}\left( x_{h+1}^{(\tau)} \right) + \zeta_{u,h}^{(k)}, \\
    \zeta_{u,h}^{(k)} &\sim \cN(0,\sigma_u^2 I_d).
\end{align*}

The value estimate $\widetilde{V}_{h+1}^{(k)}$ can change between batches, so we have to recompute the contributions of past transitions using its current value. Each new target vector receives fresh Gaussian noise, and the privacy analysis accounts for all $MH$ releases.

\textbf{Private reward estimates.} The reward functions may vary across episodes and are revealed in full only after each episode. By Definition~\ref{def:adversarial-linear}, each revealed reward function $r_h^{(j)}$ has a coefficient representation $r_h^{(j)}(x,a)=\inner{\psi(x,a)}{\theta_h^{(j)}}$ with $\norm{\theta_h^{(j)}}_2\le L$. Since $\psi$ is known and $r_h^{(j)}$ is revealed in full, we take the minimum-norm solution of the reward equations. Its norm is at most $L$ even when the reward features are rank deficient. At a batch start $k>1$, Algorithm~\ref{alg:privPO} first averages the coefficient vectors from the preceding batch:
\begin{align*}
    \bar{\theta}_h^{(k)}:=\frac{1}{B}\sum_{j=k-B}^{k-1}\theta_h^{(j)}.
\end{align*}

We then add Gaussian noise to the average coefficient vector:
\begin{align*}
    \widetilde{\theta}_h^{(k)} := \bar{\theta}_h^{(k)} + \zeta_{\theta,h}^{(k)}, \quad \zeta_{\theta,h}^{(k)}\sim \cN \left(0,\sigma_\theta^2 I_p\right).
\end{align*}
The private reward function is then reconstructed from this noisy coefficient vector and clipped to $[0,1]$:
\begin{align*}
    \widetilde{r}_h^{(k)}(x,a):=\mathsf{Clip}_{[0,1]}\left(\inner{\psi(x,a)}{\widetilde{\theta}_h^{(k)}}\right).
\end{align*}

We use this estimate to update $\widetilde{Q}_h^{(k)}$. Clipping keeps the private rewards in $[0,1]$ after noise is added. The coefficient noises are independent across batch starts and stages and independent of the other algorithm and environment randomness. We set $\bar{\theta}_h^{(1)}=0$ for the first batch. Each episode's coefficient vectors enter at most one batch average. At a fixed preceding public history, the coefficient averages over all $H$ stages have joint sensitivity at most $2L\sqrt{H}/B$.



\textbf{Exploration bonus.} In Algorithm~\ref{alg:privPO}, $z^+=\max\{z,0\}$. Limited data coverage can cause the private regression estimates above to underestimate the next-step value and discourage exploration. Following~\citet{zhong2023theoreticalanalysisoptimisticproximal}, we therefore add an exploration bonus to construct optimistic transition estimates,
\begin{align*}
    \widetilde{\Gamma}_h^{(k)}(x,a) := \beta_{\mathsf{priv}} \cdot \sqrt{\phi(x,a)^\top \left( \widetilde{\Lambda}_h^{(k)} \right)^{-1} \phi(x,a) },
\end{align*}
when $\widetilde{\Lambda}_h^{(k)} \succ 0$. Here $\beta_{\mathsf{priv}}$ is calibrated to account for both statistical error and noise in the regression statistics. We add the bonus to $\phi^\top \widetilde{w}_h^{(k)}$, clip the result to $[0, H-h]$, and combine it with the private $\widetilde{r}_h^{(k)}$ to obtain $\widetilde{Q}_h^{(k)}$. The parameter choices and the optimism guarantee are given in Appendix~\ref{sec:proof-reg}.

\begin{theorem} \label{thm:reg}
    Consider a linear MDP as defined in Definition~\ref{def:linearMDP} with rewards satisfying Definition~\ref{def:adversarial-linear}. Suppose the adversary is non-anticipating adaptive and every episode starts from the same fixed state $x_1$. Fix privacy parameters $\epsilon\in(0,1)$ and $\delta\in (0,1)$, and a failure probability $\gamma\in(0,1)$. With the noise scales specified in Lemma~\ref{lem:privacy-calibration}, there are choices of $B, \alpha, \lambda, \beta_{\mathsf{priv}}$, such that Algorithm~{\ref{alg:privPO}} is $(\epsilon, \delta)$-JDP and, with probability at least $1-\gamma$, satisfies the following regret bound~\footnote{$\widetilde{\cO}$ hides logarithmic factors. These may depend on $K,H,d,p,1+L,S,A,1/\epsilon,1/\delta$, and $1/\gamma$.} 
    \begin{align*}
        \reg(K)&\le\widetilde{\cO}\left(H^2d^{3/4}\sqrt{1+\frac{p}{d^2}}K^{3/4}+H^2d^{5/2}K^{1/2} \right. \\
        &\left.+\frac{H^{9/4}d^{5/8}K^{5/8}}{\sqrt{\epsilon}} +\frac{H^{9/4}d^{5/4}K^{1/2}}{\sqrt{\epsilon}}+\frac{H^{3/2}LK^{1/2}}{d^{3/2}\epsilon}\right).
    \end{align*}
\end{theorem}

\section{Proof Sketch}

We outline the proof of Theorem~\ref{thm:reg}, with full details in Appendix~\ref{sec:proof-reg}.

\textbf{Privacy analysis.} By the billboard lemma~\citep{hsu2016}, it is enough to prove $(\epsilon,\delta)$-DP for the common private statistics $\{\widetilde{u}, \widetilde{\Lambda}, \widetilde{r}\}$ to obtain $(\epsilon,\delta)$-JDP. The target vectors use Gaussian noise with adaptive composition of zero-concentrated differential privacy (zCDP) guarantees, the design matrix uses the binary tree mechanism, and the reward estimate uses Gaussian noise on batch-average coefficients.  Conditioning on previous private outputs fixes the value function used in each release. At a common preceding output history, fixing the adversary and the action-sampling seeds gives identical contributions from unchanged users. We combine the conditional privacy losses in the actual release order, following the zCDP analysis in Theorem~6.4 of~\citet{zhou2022differentially}. Reconstructing and clipping the reward is post-processing. The sensitivity calculation and the noise scales are given in Lemma~\ref{lem:privacy-calibration}.


\textbf{Regret analysis.} By the value difference lemma, we have~\eqref{eq:sketch-regret-decomposition}:
\begin{align}
    &\reg(K) \notag \\ &= \underbrace{\sum_{k=1}^{K} \sum_{h=1}^{H} \mathbb{E}_{\pi^\star} \left[ \inner{\widetilde{Q}_h^{(k)}(x_h,\cdot)}{\pi_h^\star(\cdot\mid x_h)-\pi_h^{(k)}(\cdot\mid x_h)}_{\cA} \right]}_{\mathsf{T}_1} \notag \\
    &+ \underbrace{\sum_{k=1}^{K} \sum_{h=1}^{H} \left(\mathbb{E}_{\pi^\star} \left[\widetilde{\delta}_h^{(k)}(x_h,a_h)\right]-\mathbb{E}_{\pi^{(k)}} \left[\widetilde{\delta}_h^{(k)}(x_h,a_h)\right]\right)}_{\mathsf{T}_2},\label{eq:sketch-regret-decomposition}
\end{align}
where $\widetilde{\delta}_h^{(k)}(x,a) := r_h^{(k)}(x,a)+\left(\mathbb{P}_h \widetilde{V}_{h+1}^{(k)} \right)(x,a) - \widetilde{Q}_h^{(k)}(x,a)$ denotes the private Bellman error.

To bound $\mathsf{T}_1$, we first note that clipping ensures $0 \le \widetilde{Q}_h^{(k)} \le H$. Following the proof of Lemma~4.2 in~\citet{zhong2023theoreticalanalysisoptimisticproximal}, we sum over batches and telescope the KL terms to obtain the bound in Lemma~\ref{lem:private-policy-error}:
\begin{align*}
    \mathsf{T}_1\le\frac{\alpha H^3K}{2}+\frac{BH\log A}{\alpha}.
\end{align*}

We now turn to $\mathsf{T}_2$. Let $t_m$ denote the first episode of batch $m$. We denote $\bar{r}_h^{(t_m)} := \frac{1}{B} \sum_{j=t_m-B}^{t_m-1}r_h^{(j)}$, for $m\ge2$, and set $\bar{r}_h^{(1)} = 0$. For an episode $k$ in this batch, the Bellman error decomposes as
\begin{align}
    \widetilde{\delta}_h^{(k)}&=\underbrace{r_h^{(k)}-\bar{r}_h^{(t_m)}}_{\text{reward mismatch}}+\underbrace{\bar{r}_h^{(t_m)}-\widetilde{r}_h^{(t_m)}}_{\text{reward noise}} \notag \\ 
    &\qquad +\underbrace{\mathbb{P}_h\widetilde{V}_{h+1}^{(t_m)}-\widehat{\mathbb{P}}_h^{(t_m)}\widetilde{V}_{h+1}^{(t_m)}}_{\text{transition error}}.\label{eq:sketch-bellman-decomposition}
\end{align}
In~\eqref{eq:sketch-bellman-decomposition}, the mismatch between each episode's reward and the previous batch's average reward contributes at most $BH+\alpha H^3K$. Clipping cannot increase the distance between a noisy reward and the batch average, so Gaussian concentration bounds the contribution from reward noise by $2KH\bar{\zeta}_r$, where $\bar{\zeta}_r$ denotes a uniform bound on the difference between $\widetilde{r}_h^{(t_m)}$ and $\bar{r}_h^{(t_m)}$. Its choice and the corresponding noise bounds are given in Lemma~\ref{lem:private-noise-bounds}. 



\begin{remark}[Reward noise and high-probability regret]
    One may ask why our regret guarantee holds with high probability, while the noise-injection argument of~\citet{agarwal2017price} is in expectation. Their argument equates the expected cumulative losses of the private algorithm and an auxiliary algorithm with a single initial perturbation (their Eq.~(1)), but this equality does not by itself give a high-probability guarantee. Our proof does not rely on this identity. Instead, Lemma~\ref{lem:private-noise-bounds} bounds the reward errors uniformly with high probability, giving the $2KH\bar{\zeta}_r$ term without averaging over the noise. Although this bound appears linear in $K$, our choice of batch size makes the reward-noise contribution sublinear; see~\eqref{eq:sketch-regret-before-tuning} and Theorem~\ref{thm:reg}.
\end{remark}

The transition error requires more care because noise affects both the regression statistics and the rewards used to construct the value functions. We condition on the independent coefficient noise and construct a value function class containing every sequence of bounded reward coefficients, together with the regression weights and bonus matrices. This class is fixed conditional on the noise, even though the realized reward coefficients remain adaptive. Covering the coefficients adds $pM$ parameters to the original $d^2M$ count. We bound the statistical error using the covering argument in the proof of Lemma~B.3 in~\citet{zhong2023theoreticalanalysisoptimisticproximal}. Let $\Sigma_E$ and $\Sigma_u$ denote uniform bounds on the spectral norm of the matrix noise and the Euclidean norm of the target-vector noise, respectively, with their choices given in Lemma~\ref{lem:private-noise-bounds}. For $\lambda\ge\max\left\{1,2\Sigma_E\right\}$, we have the confidence radius as in Lemma~\ref{lem:private-confidence}:
\begin{align*}
    \beta_{\mathsf{priv}}=\widetilde{\cO}\left(H\sqrt{(d^2+p) M}+\frac{\Sigma_u}{\sqrt{\lambda}}+H\sqrt{\lambda d}\right).
\end{align*}

The three terms account for statistical error, target-vector noise, and regularization, respectively. The matrix noise is controlled by choosing $\lambda\ge2\Sigma_E$. Adding the bonus with this confidence radius makes the transition estimate optimistic, and clipping preserves this property (Lemma~\ref{lem:private-transition-optimism}). The transition-error term under $\pi^\star$ is therefore non-positive. The learner's expected transition error is bounded by the sum of errors along the observed trajectories plus a martingale deviation term obtained by the same Azuma-Hoeffding argument as in the proof of Lemma~4.3 of~\citet{zhong2023theoreticalanalysisoptimisticproximal}, since the current estimates are fixed before each episode's trajectory is sampled. Each error along a trajectory is at most twice the clipped bonus. Including the reward terms gives~\eqref{eq:sketch-bellman-bonus-bound}:
\begin{align}
    \mathsf{T}_2& \le 2\sum_{m=1}^{M} \sum_{k=t_m}^{t_{m+1}-1}\sum_{h=1}^{H}\min\left\{H,\widetilde{\Gamma}_h^{(t_m)}\left(x_h^{(k)},a_h^{(k)}\right)\right\} \notag \\
    &+BH+\alpha H^3K+ 3H^{3/2} \sqrt{2K\log\left(\frac{8}{\gamma}\right)}+2KH\bar{\zeta}_r.\label{eq:sketch-bellman-bonus-bound}
\end{align}

These bonuses remain fixed within each batch, even as new data arrive. Following~\citet{chowdhury2022shuffle}, we compare them with bonuses computed from virtual design matrices that include all earlier episodes and contain no noise. The determinant argument in Lemma~\ref{lem:private-bonus-sum} gives
\begin{align*}
    &\sum_{m=1}^{M}\sum_{k=t_m}^{t_{m+1}-1}\sum_{h=1}^{H}\min\left\{H,\widetilde{\Gamma}_h^{(t_m)}\left(x_h^{(k)},a_h^{(k)}\right)\right\}\\
    &\qquad \le\widetilde{\cO}\left(H^2dB+H\beta_{\mathsf{priv}}\sqrt{dK}\right),
\end{align*}
where $H^2dB$ is the cost of keeping the bonuses fixed within each batch.

Finally, we combine the bounds for $\mathsf{T}_1$ and $\mathsf{T}_2$. With $\alpha=H^{-1}\sqrt{2B\log A/(3K)}$, substituting $\beta_{\mathsf{priv}}$ and the noise bounds from Lemma~\ref{lem:private-noise-bounds} gives~\eqref{eq:sketch-regret-before-tuning}:
\begin{align}
    \reg(K)&\le\widetilde{\cO}\left(\underbrace{H^2\sqrt{BK}+\frac{H^2 \sqrt{d (d^2+p)} K}{\sqrt{B}}}_{(a)}+\underbrace{H^2dB}_{(b)}\right. \notag \\
    &\qquad\left.+\underbrace{H^2d\sqrt{\lambda K}+\frac{H^{5/2}dK}{\epsilon\sqrt{B\lambda}}+\frac{H^{3/2} LK}{B\epsilon}}_{(c)}\right).\label{eq:sketch-regret-before-tuning}
\end{align}
Here $(a)$ contains the policy optimization and statistical transition errors, $(b)$ is the cost of batch updates in the bonus sum, and $(c)$ contains the regularization and noise terms. For $K \ge d^3$, the choices $B=\sqrt{d^3K}$ and $\lambda=\max\left\{1,2\Sigma_E,\Sigma_u/(H\sqrt{d})\right\}$ give the regret bound in Theorem~\ref{thm:reg} with probability at least $1-\gamma$. When $K<d^3$, we take $B=K$, and the trivial bound $\reg(K)\le HK\le H^2d^{3/4}K^{3/4}$ implies the same regret guarantee. The full proof is given in Appendix~\ref{sec:proof-reg}.

\begin{remark}[Effect of batching and privacy]
    The two terms in $(a)$ show how the batch size affects regret: larger batches increase the policy optimization error but reduce the statistical transition error. For $K \ge d(d^2+p)$, these two terms balance at  $B\asymp\sqrt{d\left(d^2+p\right)K}$. The batch size in Theorem~\ref{thm:reg} matches the order used in~\citet{zhong2023theoreticalanalysisoptimisticproximal} and, when $K\ge d^3$, gives $M=K/B=\sqrt{K}/d^{3/2}$ updates. For fixed $d,p,L,H,S,A,\epsilon,\delta$, and $\gamma$, the privacy terms in Theorem~\ref{thm:reg} grow as $K^{5/8}$ and $K^{1/2}$, up to logarithmic factors. Thus, privacy does not change the leading $K^{3/4}$ dependence of this bound. The specific choice of parameter is given in Appendix~\ref{sec:proof-reg}.
\end{remark}

\begin{remark}[Privacy costs]
    The noise scales depend on how many releases one user can affect. For the target vectors, zCDP composition over at most $MH$ Gaussian releases gives the $\sqrt{M}$ factor in $\sigma_u$. The binary tree mechanism limits each user's contribution to at most $1+ \log_2 K$ nodes per stage, avoiding this factor for the design matrices. Each user's rewards appear in only one batch average, so the reward releases have no composition cost across batches, and their sensitivity is proportional to $1/B$. These noise scales are derived in Lemma~\ref{lem:privacy-calibration}.
\end{remark}

\begin{remark}[Differences from the non-private analysis]
    Two parts of the transition analysis need to account for privacy noise. First, the estimated value functions depend on the noisy rewards. The adaptive reward coefficients are included in the value function class. Conditioning on the independent coefficient noise leaves a fixed class with log-covering number of order $(d^2+p)M$, up to logarithms, without introducing a polynomial $SA$ factor; see Lemma~\ref{lem:private-confidence}. Second, the determinant argument relies on design matrices that increase in positive semi-definite order as data are added. The noisy matrices need not have this property. We therefore apply the argument to virtual design matrices, which contain no noise and increase as data are added. Comparing their bonuses with the private bonuses using the noise bounds then gives the $H^2dB$ cost of batch updates. The comparison and determinant calculation are given in Lemma~\ref{lem:private-bonus-sum}.
\end{remark}
\section{Conclusion}

We establish a high-probability JDP regret guarantee for policy optimization with unknown linear transitions and adaptive adversarial linear rewards under full-information feedback. The algorithm protects both trajectories and reward functions. Batching limits releases of regression target vectors, the binary tree mechanism protects covariance matrices, and batch averaging reduces the sensitivity of reward coefficients. Our analysis accounts for adaptive reward choices in both privacy composition and concentration. For fixed model and privacy parameters, the leading regret remains $\widetilde{\cO}(K^{3/4})$, with additive privacy terms of $\widetilde{\cO}(K^{5/8})$ and $\widetilde{\cO}(K^{1/2})$. Extending these guarantees to bandit feedback and improving the leading regret rate under JDP are directions for future work.

\section{Open Problem}

Consider the same unknown linear transitions, but allow each episode to have arbitrary bounded reward functions. In an earlier analysis with the reward sequence fixed before the interaction, we obtained a $\widetilde{\cO}(K^{3/4})$ regret bound for fixed model and privacy parameters. The dependence on the state space size $S$ arises because changing one user can change all $SA$ entries of a reward table. With batch size $B=d^{3/2}\sqrt{K}$, adding Gaussian noise to the batch averages contributes $\widetilde{\cO}\!\left(\frac{H^{3/2}\sqrt{SAK}}{d^{3/2}\epsilon}\right)$ to the regret bound.

\emph{With unknown linear transitions and arbitrary bounded rewards, can policy optimization under full-information feedback and episode-level JDP achieve a regret bound that is sublinear in $K$ and independent of $S$?}

\setcounter{biburllcpenalty}{200}
\printbibliography
\newpage

\appendix
\renewcommand{\contentsname}{Contents of Appendix}
\addtocontents{toc}{\protect\setcounter{tocdepth}{2}}

{
  \tableofcontents
}

\section{Detailed Related Work} \label{sec:related}

\textbf{Adversarial MDPs.} In this paper, we consider adversarial MDPs, where the reward functions are chosen by a non-anticipating adaptive adversary before each episode begins and may differ arbitrarily across episodes. Earlier adversarial MDP results cover several different transition and feedback settings. 

\emph{Known transition and full reward information.}\citet{Yu2009MarkovDP, even2009online} first studied the known transition probability and full reward information setting, achieving $\cO(K^{2/3})$ and $\cO(\sqrt{K})$, respectively.~\footnote{In this section, $K$ denotes the number of interaction steps for continuing MDPs and the number of episodes for episodic MDPs.}

\emph{Known transition and bandit reward feedback.} In continuing MDPs,~\citet{neu2010} proposed the MDP-EXP3 algorithm and obtained $\widetilde{O}\left(K^{2/3}\right)$ regret. For loop-free episodic MDPs,~\citet{neu2010online} achieved an improved regret bound of $\cO\left(H^2\sqrt{AK}/\alpha\right)$ under the assumption that all states are reachable with a probability at least $\alpha>0$ under every policy. Furthermore,~\citet{zimin2013online} introduced the O-REPS algorithm, removing the reachability assumption and obtaining $\widetilde{O}(\sqrt{K})$ regret. \citet{neu2021online} obtain $\widetilde{\cO}(\sqrt{K})$ expected regret when rewards and transitions are linear in the same features. Their result assumes that the exploration policy induces a non-degenerate feature covariance matrix.

\emph{Unknown transition and full reward information.} For tabular MDPs,~\citet{neu12} proposed the FPOP algorithm that achieves $\widetilde{O} (HSA\sqrt{K})$ regret,~\citet{rosenberg2019onlineconvexoptimizationadversarial} then improved the regret to $\widetilde{O}(HS\sqrt{AK})$.  \citet{cai2020provablyefficientexplorationpolicy} study linear mixture MDPs and obtain $\widetilde{O}(d H^{2} \sqrt{K})$ regret, and this result is further improved to $\widetilde{O}(d H^{3/2} \sqrt{K})$ by~\citet{he2022nearoptimalpolicyoptimizationalgorithms}. \citet{ji2024horizon} obtained the first horizon-free regret bound for adversarial linear mixture MDPs with rewards in $[0,1/H]$; rescaling to rewards in $[0,1]$ gives $\widetilde{\cO} (Hd\sqrt{K}+Hd^2)$. ~\citet{zhong2023theoreticalanalysisoptimisticproximal} developed an optimistic variant of PPO that achieves a regret bound of $\widetilde{\cO}\left(d^{3/4}H^2K^{3/4} + d^{5/2}H^2K^{1/2} \right)$ in linear MDPs without requiring linear rewards. When the rewards are linear in the transition feature,~\citet{sherman2024rateoptimalpolicyoptimizationlinear} obtain $\widetilde{\cO}(\sqrt{K})$ regret with a computationally efficient policy optimization algorithm against oblivious adversaries, after a reward-free warm-up phase.~\citet{cassel2024warm} retain this rate while removing the warm-up phase through feature contraction, improving the dependence on $H$ and $d$, and allowing adaptive adversaries. For comparison, in stochastic MDPs with fixed expected rewards,~\citet{liu2023optimistic} proposed Optimistic NPG, which learns an $\varepsilon_{\mathsf{opt}}$-optimal policy in a $d$-dimensional linear MDP using $\widetilde{\cO}(d^2 H^6/\varepsilon_{\mathsf{opt}}^3)$ episodes and achieves polynomial sample complexity under general function approximation.

\emph{Unknown transition and bandit reward feedback.} ~\citet{rosenberg2019online} achieve a regret bound of $\widetilde{O}(K^{3/4})$ and improve this to $\widetilde{O}(\sqrt{K}/\alpha)$ regret under the assumption that every state is reachable with probability at least $\alpha$. Later, ~\citet{jin2020learning} achieve $\widetilde{O}(\sqrt{K})$ regret without the additional assumption.~\citet{luo2021policyoptimizationadversarialmdps} study bandit-feedback linear-$Q$ MDPs with a simulator, providing a regret bound of $\widetilde{O}(d^{2/3} H^2 K^{2/3})$. They also study simulator-free bandit-feedback linear MDPs, obtaining $\widetilde{O}(d^{2} H^4 K^{14/15})$ regret.~\citet{dai2023refined} further improve these regret to nearly optimal $\widetilde{O}(\sqrt{K})$ with a simulator and $\widetilde{O}( K^{8/9})$ without a simulator. ~\citet{sherman2023improvedregretefficientonline} also improved the bound to $\widetilde{O}(K^{2/3})$ with weaker simulator and $\widetilde{O}(K^{6/7})$ without simulator. With rewards linear in the transition feature,~\citet{kong2024improvedregretboundslinear} obtain $\widetilde{\cO}(K^{4/5})$ regret plus an additive term polynomial in the inverse of a minimum-eigenvalue parameter, using a computationally inefficient algorithm without a simulator.~\citet{liu2024towards} remove this dependence and obtain $\widetilde{\cO}(\sqrt{K})$ expected regret with a computationally inefficient algorithm and $\widetilde{\cO}(K^{3/4})$ expected regret with a computationally efficient policy optimization algorithm, against oblivious adversaries and without a simulator.
 
\textbf{Private adversarial MDPs.} \citet{bai2024differentially} study tabular loop-free adversarial MDPs with unknown transitions against an oblivious adversary under JDP and LDP, obtaining expected regret guarantees under full information and bandit feedback. Their full-information JDP bound has an additive privacy cost independent of $K$ up to logarithmic factors, using the noise-injection approach of~\citet{agarwal2017price} for private online linear optimization. For adversarial bandits with $A$ arms over $K$ rounds,~\citet{asi2025faster} obtain $\cO\left(\sqrt{AK/\epsilon}\right)$ expected regret under central $\epsilon$-DP for $\epsilon\in[1/K,1]$, separating central DP from LDP. Using a lazy-to-private transformation,~\citet{asi2024private} obtain expected regret bounds of $\widetilde{\cO}\left(\sqrt{K\log N}+K^{1/3}\log N/\epsilon^{2/3}\right)$ for full-information prediction with $N$ experts and $\widetilde{\cO}\left(\sqrt{K}+K^{1/3}\sqrt{d_{\mathsf{oco}}}/\epsilon^{2/3}\right)$ for $d_{\mathsf{oco}}$-dimensional online convex optimization under $(\epsilon,\delta)$-DP, with unit Lipschitz constant and domain diameter in the latter setting. While these bounds have a smaller leading dependence on $K$ for fixed dimensions and privacy parameters, they concern settings without unknown transition dynamics and provide guarantees in expectation against oblivious adversaries.

\section{Proof of Theorem~\ref{thm:reg}} \label{sec:proof-reg}

We begin with a standard regret decomposition. Using the best fixed policy $\pi^{\star}$ defined in the preliminaries, we write
\begin{align*}
    \reg(K) 
    &= \sum_{k=1}^K \left( V_1^{(\pi^\star, k)}(x_1^{(k)}) -  V_1^{(\pi^{(k)}, k)}(x_1^{(k)})\right) \\
    &= \sum_{k=1}^K \left( \underbrace{V_1^{(\pi^\star, k)}(x_1^{(k)}) - \widetilde{V}_1^{(k)}(x_1^{(k)})}_{(i)} + \underbrace{\widetilde{V}_1^{(k)}(x_1^{(k)}) - V_1^{(\pi^{(k)}, k)}(x_1^{(k)})}_{(ii)}\right).
\end{align*}

We restate the value difference identity from~\citet[Appendix B.1]{cai2020provablyefficientexplorationpolicy} in our notation, following the formulation in Lemma~B.1 of~\citet{zhong2023theoreticalanalysisoptimisticproximal}. 

\begin{lemma}[Value difference lemma]
\label{lem:value-difference}
Fix an episode $k \in [K]$, a policy $\pi^{(k)}$, and bounded functions $\{\widetilde{Q}_h^{(k)}:\cS \times \cA \rightarrow \mathbb{R} \}_{h=1}^{H}$. Let $\widetilde{V}_h^{(k)}(x) := \inner{\widetilde{Q}_h^{(k)}(x,\cdot)}{\pi_h^{(k)}(\cdot | x)}_{\cA}$. Set $\widetilde{V}_{H+1}^{(k)}\equiv 0$. Then, for any policy $\pi'$, we have
\begin{align*}
    \widetilde{V}_1^{(k)} \left(x_1^{(k)}\right) - V_1^{(\pi',k)} \left(x_1^{(k)}\right)
    &= \sum_{h=1}^{H} \mathbb{E}_{\pi'}\left[\inner{\widetilde{Q}_h^{(k)} (x_h,\cdot)}{\pi_h^{(k)}(\cdot | x_h) - \pi'_h(\cdot | x_h)}_{\cA}\right] \\
    &\quad + \sum_{h=1}^{H}\mathbb{E}_{\pi'}\left[\widetilde{Q}_h^{(k)}(x_h,a_h)-r_h^{(k)}(x_h,a_h)-\left(\mathbb{P}_h\widetilde{V}_{h+1}^{(k)}\right)(x_h,a_h)\right].
\end{align*}
\end{lemma}

For the first term $(i)$, applying Lemma~\ref{lem:value-difference} by setting $\pi' = \pi^\star$, we have
\begin{align*}
    \widetilde{V}_1^{(k)} \left(x_1^{(k)}\right) - V_1^{(\pi^\star, k)} \left(x_1^{(k)}\right) 
    &= \sum_{h=1}^H \mathbb{E}_{\pi^\star} \left[ \inner{\widetilde{Q}_h^{(k)} (x_h,\cdot)}{\pi_h^{(k)}(\cdot | x_h) - \pi_h^{\star} (\cdot | x_h)}_{\cA} \right] \\
    &\quad + \sum_{h=1}^{H} \mathbb{E}_{\pi^\star} \left[ \widetilde{Q}_h^{(k)}(x_h,a_h) - \left( r_h^{(k)}(x_h,a_h) + \left(\mathbb{P}_h \widetilde{V}_{h+1}^{(k)} \right) (x_h,a_h) \right) \right],
\end{align*}
thus we have
\begin{align*}
    (i) = \sum_{h=1}^H \mathbb{E}_{\pi^\star} \left[ \inner{\widetilde{Q}_h^{(k)}(x_h,\cdot)}{ \pi_h^{\star}(\cdot |  x_h) - \pi_h^{(k)}(\cdot | x_h) }_{\cA} \right] + \sum_{h=1}^H \mathbb{E}_{\pi^\star}[\widetilde{\delta}_h^{(k)}(x_h,a_h)],
\end{align*}
where we define $\widetilde{\delta}_h^{(k)}(x_h,a_h) = \left( r_h^{(k)}(x_h,a_h) + \left(\mathbb{P}_h \widetilde{V}_{h+1}^{(k)}\right)(x_h,a_h) \right) - \widetilde{Q}_h^{(k)}(x_h,a_h)$ as the private Bellman error.

Similarly, applying the same lemma to term $(ii)$ with $\pi' = \pi^{(k)}$, yield
\begin{align*}
    (ii) &= \sum_{h=1}^{H} \mathbb{E}_{\pi^{(k)}} \left[ \widetilde{Q}_h^{(k)}(x_h,a_h) - \left( r_h^{(k)}(x_h,a_h) + \left(\mathbb{P}_h \widetilde{V}_{h+1}^{(k)}\right)(x_h,a_h) \right) \right] \\
    &= - \sum_{h=1}^{H} \mathbb{E}_{\pi^{(k)}} \left[ \widetilde{\delta}_h^{(k)}(x_h,a_h) \right].
\end{align*}

Thus, combining $(i)$ and $(ii)$ above, the total regret can be decomposed as
\begin{align} 
    \reg(K) &= \sum_{k=1}^K \sum_{h=1}^H \mathbb{E}_{\pi^\star} \left[ \inner{\widetilde{Q}_h^{(k)}(x_h,\cdot)}{ \pi_h^{\star}(\cdot |  x_h) - \pi_h^{(k)}(\cdot |  x_h) }_{\cA} \right] \notag \\
    &\quad + \sum_{k=1}^K \sum_{h=1}^H \left( \mathbb{E}_{\pi^\star}[\widetilde{\delta}_h^{(k)}(x_h,a_h)] - \mathbb{E}_{\pi^{(k)}} \left[ \widetilde{\delta}_h^{(k)}(x_h,a_h) \right] \right). \label{eq:private-regret-decomposition}
\end{align}

Let $M:=K/B$ denote the number of batches, and $t_m := (m-1)B+1$ denote the first episode of batch $m\in[M]$, with $t_{M+1}:=K+1$. Episode $k$ belongs to batch $m$ if $t_m\le k<t_{m+1}$.

We first bound the policy optimization error using the same argument as in Lemma~4.2 of \citet{zhong2023theoreticalanalysisoptimisticproximal}.

\begin{lemma}[Policy optimization error] \label{lem:private-policy-error}
    Suppose every episode starts from the same fixed state $x_1$. Then
    \begin{align*}
        \sum_{k=1}^{K} \sum_{h=1}^{H} \mathbb{E}_{\pi^\star}\left[ \inner{\widetilde{Q}_h^{(k)}(x_h,\cdot)}{\pi_h^\star (\cdot\mid x_h) - \pi_h^{(k)}(\cdot\mid x_h)}_{\cA}\right] \le \frac{\alpha H^3K}{2} + \frac{BH\log A}{\alpha}.
    \end{align*}
\end{lemma}

\begin{proof}
    Recall that we clip $\widetilde{r}_h^{(t_m)}$ to $[0,1]$ and $\widehat{\mathbb{P}}_h^{(t_m)}\widetilde{V}_{h+1}^{(t_m)}$ to $[0,H-h]$. Thus, $0 \le \widetilde{Q}_h^{(t_m)}(x,a) \le H$ for every $(m,h,x,a)$, so the boundedness condition used in the policy-update analysis of~\citet{zhong2023theoreticalanalysisoptimisticproximal} still holds for our private estimates.

    Fix $h\in[H]$. Applying the KL telescoping argument in Appendix~B.2 of \citet{zhong2023theoreticalanalysisoptimisticproximal}, specifically (B.10)~\footnote{For the last batch, the next policy in the KL telescoping argument is defined by the same update solely for analysis.}, we obtain, for every state $x$,
    \begin{align*}
        \sum_{m=1}^{M}\inner{\widetilde{Q}_h^{(t_m)}(x,\cdot)}{\pi_h^\star(\cdot\mid x)-\pi_h^{(t_m)}(\cdot\mid x)}_{\cA} \le \frac{\alpha H^2 M}{2}+\frac{\log A}{\alpha}.
    \end{align*}
    
   Since the initial state and transition kernel are the same in every episode, the distribution of $x_h$ under $\pi^\star$ is also the same. Taking $\mathbb{E}_{\pi^\star}$, multiplying by $B$, and using the fact that the policy and $Q$-function remain unchanged within each batch, we obtain
    \begin{align*}
        \sum_{k=1}^{K}\mathbb{E}_{\pi^\star}\left[\inner{\widetilde{Q}_h^{(k)}(x_h,\cdot)}{\pi_h^\star(\cdot\mid x_h)-\pi_h^{(k)}(\cdot\mid x_h)}_{\cA}\right] \le \frac{\alpha H^2K}{2}+\frac{B\log A}{\alpha}.
    \end{align*}
    
    Summing over $h\in[H]$ proves the lemma.
    
\end{proof}

\begin{lemma} 
    It holds with probability at least $1-2\gamma$ that
    \begin{align}
        &\sum_{k=1}^K \sum_{h=1}^H \left( \mathbb{E}_{\pi^\star}[\widetilde{\delta}_h^{(k)}(x_h,a_h)] - \mathbb{E}_{\pi^{(k)}}[\widetilde{\delta}_h^{(k)}(x_h,a_h)] \right) \notag\\
        &\quad \le \sum_{m=1}^M \sum_{k=t_m}^{t_{m+1}-1} \sum_{h=1}^H \mathbb{E}_{\pi^\star} \left[ \left( \mathbb{P}_h \widetilde{V}_{h+1}^{(t_m)} - \widehat{\mathbb{P}}_h^{(t_m)} \widetilde{V}_{h+1}^{(t_m)} \right) (x_h,a_h) \right] \notag \\
        &\qquad + \sum_{m=1}^M \sum_{k=t_m}^{t_{m+1}-1} \sum_{h=1}^H \left( \widehat{\mathbb{P}}_h^{(t_m)} \widetilde{V}_{h+1}^{(t_m)} - \mathbb{P}_h \widetilde{V}_{h+1}^{(t_m)} \right) (x_h^{(k)},a_h^{(k)}) \notag \\
        &\qquad + BH + \alpha H^3 K + 3\sqrt{2H^3K\log\left(\frac{2}{\gamma}\right)} + 2KH\bar{\zeta}_r. \label{eq:private-bellman}
    \end{align}
\end{lemma}

\begin{proof}

    For each $h\in[H]$, we define the average reward as $\bar{r}_h^{(t_m)}:=B^{-1}\sum_{j=t_m-B}^{t_m-1}r_h^{(j)}$ for $m\ge2$, and set $\bar{r}_h^{(t_1)}=0$. For every episode $k$ in batch $m$, we have $\widetilde{Q}_h^{(k)} = \widetilde{Q}_h^{(t_m)} = \widetilde{r}_h^{(t_m)} + \widehat{\mathbb{P}}_h^{(t_m)} \widetilde{V}_{h+1}^{(t_m)}$.
    Hence,
    \begin{align*}
        \widetilde{\delta}_{h}^{(k)}(x_h,a_h)
        &= \left( r_{h}^{(k)}(x_h,a_h) - \widetilde{r}_h^{(t_m)}(x_h,a_h) \right)
        + \left( \mathbb{P}_{h}\widetilde{V}_{h+1}^{(t_m)} - \widehat{\mathbb{P}}_{h}^{(t_m)}\widetilde{V}_{h+1}^{(t_m)} \right)(x_h,a_h) \\
        &= \underbrace{\left( r_{h}^{(k)}(x_h,a_h) - \bar{r}_h^{(t_m)}(x_h,a_h) \right)}_{(i)} + \underbrace{\left( \bar{r}_h^{(t_m)}(x_h,a_h) - \widetilde{r}_h^{(t_m)}(x_h,a_h) \right)}_{(ii)} \\
        &\quad + \underbrace{\left( \mathbb{P}_{h}\widetilde{V}_{h+1}^{(t_m)} - \widehat{\mathbb{P}}_{h}^{(t_m)}\widetilde{V}_{h+1}^{(t_m)} \right)(x_h,a_h)}_{(iii)}.
    \end{align*}

    We first consider term $(i)$. Following the same reward-mismatch analysis as in Lemma~4.3 of~\citet{zhong2023theoreticalanalysisoptimisticproximal}, for each $h\in[H]$, we have
    \begin{align*}
        \sum_{m=1}^M \sum_{k=t_m}^{t_{m+1}-1} \mathbb{E}_{\pi^\star}\left[ r_h^{(k)}(x_h,a_h)-\bar{r}_h^{(t_m)}(x_h,a_h)
        \right] &\le B, \\
        \sum_{m=1}^M \sum_{k=t_m}^{t_{m+1}-1} \mathbb{E}_{\pi^{(k)}}\left[ \bar{r}_h^{(t_m)}(x_h,a_h)-r_h^{(k)}(x_h,a_h)
        \right] &\le \alpha H^2K.
    \end{align*}
    
    The same argument applies because the private estimates satisfy $0 \le \widetilde{Q}_h^{(k)}(x,a)\le H$. For the second inequality, we apply the value difference lemma with the reward supported only at stage $h$. Summing over $h\in[H]$ yields
    \begin{align*}
        \sum_{m=1}^M \sum_{k=t_m}^{t_{m+1}-1} \sum_{h=1}^H \left(\mathbb{E}_{\pi^\star}-\mathbb{E}_{\pi^{(k)}}\right)
        \left[ r_h^{(k)}(x_h,a_h)-\bar{r}_h^{(t_m)}(x_h,a_h) \right] \le BH + \alpha H^3K.
    \end{align*}

    We next consider term $(ii)$. Define the reward perturbation $\zeta_{r,h}^{(t_m)}(x,a):=\inner{\psi(x,a)}{\zeta_{\theta,h}^{(t_m)}}$. Recall that $\bar{r}_h^{(t_m)}(x,a)=\inner{\psi(x,a)}{\bar{\theta}_h^{(t_m)}}$ and $\widetilde{r}_h^{(t_m)} = \mathsf{Clip}_{[0,1]}\left(\bar{r}_h^{(t_m)}+\zeta_{r,h}^{(t_m)}\right)$. Since $\bar{r}_h^{(t_m)}(x,a)\in[0,1]$ and projection onto $[0,1]$ is non-expansive, for every $(x,a)$,
    \begin{align*}
        \abs{\bar{r}_h^{(t_m)}(x,a) - \widetilde{r}_h^{(t_m)}(x,a)} \le \abs{\zeta_{r,h}^{(t_m)}(x,a)}.
    \end{align*}

    By a Gaussian tail bound and a union bound over all batch starts, stages, and state-action pairs, with probability at least $1-\gamma$, the clipping inequality above gives
    \begin{align*}
    \norm{\bar{r}_h^{(t_m)} -\widetilde{r}_h^{(t_m)}}_{\infty}
    \le \bar{\zeta}_r
    \end{align*}
    simultaneously for all $m\in[M]$ and $h\in[H]$. The choice of $\bar{\zeta}_r$ is given in Lemma~\ref{lem:private-noise-bounds}.
    
    Consequently, on the same event,
    \begin{align*}
    \mathbb{E}_{\pi^\star} \left[\bar{r}_h^{(t_m)}(x_h,a_h)-\widetilde{r}_h^{(t_m)}(x_h,a_h)\right]
    - \mathbb{E}_{\pi^{(k)}} \left[\bar{r}_h^{(t_m)}(x_h,a_h)-\widetilde{r}_h^{(t_m)}(x_h,a_h)\right]
    \le 2\bar{\zeta}_r.
    \end{align*}

    Summing over batches, episodes, and stages gives
    \begin{align*}
        \sum_{m=1}^{M}\sum_{k=t_m}^{t_{m+1}-1} \sum_{h=1}^{H} \left( \mathbb{E}_{\pi^\star} - \mathbb{E}_{\pi^{(k)}} \right) \left[ \bar{r}_h^{(t_m)}(x_h,a_h) - \widetilde{r}_h^{(t_m)}(x_h,a_h) \right] \le 2KH\bar{\zeta}_r.
    \end{align*}
    
    For term $(iii)$, we apply the same Azuma--Hoeffding argument as in the proof of Lemma~4.3 of~\citet{zhong2023theoreticalanalysisoptimisticproximal}, with probability at least $1-\gamma$,
    \begin{align*}
        &-\sum_{m=1}^{M}\sum_{k=t_m}^{t_{m+1}-1}\sum_{h=1}^{H}\mathbb{E}_{\pi^{(k)}}\left[\left(\mathbb{P}_h\widetilde{V}_{h+1}^{(t_m)}-\widehat{\mathbb{P}}_h^{(t_m)}\widetilde{V}_{h+1}^{(t_m)}\right)(x_h,a_h)\right] \\
        &\quad\le\sum_{m=1}^{M}\sum_{k=t_m}^{t_{m+1}-1}\sum_{h=1}^{H}\left(\widehat{\mathbb{P}}_h^{(t_m)}\widetilde{V}_{h+1}^{(t_m)}-\mathbb{P}_h\widetilde{V}_{h+1}^{(t_m)}\right)\left(x_h^{(k)},a_h^{(k)}\right) \\
        &\qquad+3\sqrt{2H^3K\log\left(\frac{2}{\gamma}\right)}.
    \end{align*}
    
    Combining the bounds for terms $(i)$, $(ii)$, and $(iii)$ and applying a union bound yields the stated inequality with probability at least $1-2\gamma$.
    
\end{proof}

It remains to bound the transition-estimation terms. To this end, we first establish a uniform confidence bound for the private least-squares estimator. For each episode $\tau$, let $\phi_h^{(\tau)} := \phi\left( x_h^{(\tau)}, a_h^{(\tau)} \right)$. Recall that
\begin{align*}
    \Lambda_h^{(t_m)} := \sum_{\tau = 1}^{t_m - 1} \phi_h^{(\tau)} \left( \phi_h^{(\tau)} \right)^\top, 
    \quad \widetilde{\Lambda}_h^{(t_m)} := \Lambda_h^{(t_m)} + \lambda I_d + E_h^{(t_m)}.
\end{align*}


Let $\cE_{\mathsf{noise}}$ denote the event that
\begin{align*}
    \norm{E_h^{(t_m)}}_{\mathsf{op}} \le \Sigma_E,
    \qquad \norm{\zeta_{u,h}^{(t_m)}}_2 \le \Sigma_u,
\end{align*}
simultaneously for all $m \in [M]$ and $h \in [H]$. By Gaussian concentration and a union bound, the deterministic bounds $\Sigma_E$ and $\Sigma_u$ can be chosen so that $\mathbb{P}\left[\cE_{\mathsf{noise}}\right] \ge 1-\gamma$; their explicit expressions will be specified later. We choose $\lambda \ge \max\{1,2\Sigma_E\}$, which ensures that $\norm{E_h^{(t_m)}}_{\mathsf{op}} \le \lambda/2$ on $\cE_{\mathsf{noise}}$.

\begin{lemma} \label{lem:private-confidence}
    Take $\lambda\ge\max\{1,2\Sigma_E\}$ and use the noise bounds of Lemma~\ref{lem:private-noise-bounds}. With probability at least $1-2\gamma$, simultaneously for all $m \in [M]$, $h \in [H]$, and $(x,a) \in \cS \times \cA$,
    \begin{align*}
        \abs{\phi(x,a)^\top \widetilde{w}_h^{(t_m)} - \left( \mathbb{P}_h   \widetilde{V}_{h+1}^{(t_m)} \right) (x,a)} \le \beta_{\mathsf{priv}} \sqrt{\phi(x,a)^\top \left( \widetilde{\Lambda}_h^{(t_m)} \right)^{-1} \phi(x,a) },
    \end{align*}
    where, for a sufficiently large absolute constant $c_1 \ge 2$,
    \begin{align*}
        \iota &:= \log \left( \frac{c_1(d^2+p)HK(1+\alpha)(1+L)(1+\Sigma_u)}{\gamma} \right), \\
        \beta_{\mathsf{priv}} &:= c_1 H\sqrt{(d^2+p) M \iota } + \Sigma_u \sqrt{\frac{2}{\lambda}} + H \sqrt{\frac{3\lambda d}{2}}.
    \end{align*}
\end{lemma}

\begin{proof}
    Fix $(m,h)$. For simplicity, we write $\phi=\phi(x,a)$, $\widetilde{V}:=\widetilde{V}_{h+1}^{(t_m)}$, and $\widetilde{\Lambda}:=\widetilde{\Lambda}_h^{(t_m)}$. Define $\bH:=\lambda I_d+E_h^{(t_m)}$ and $z:=\inner{\mu_h}{\widetilde{V}}_{\cS}$. The linear representation gives $(\mathbb{P}_h \widetilde{V})(x,a)=\phi^\top z$. Since clipping ensures $0\le \widetilde{V} \le H$, definition~\ref{def:linearMDP} gives $\norm{z}_2\le H\sqrt{d}$. By the definition of $\widetilde{w}_h^{(t_m)}$, we have
    \begin{align*}
        \phi^\top \widetilde{w}_h^{(t_m)} - (\mathbb{P}_h \widetilde{V})(x,a) &=\underbrace{\phi^\top \widetilde{\Lambda}^{-1} \sum_{\tau=1}^{t_m-1} \phi_h^{(\tau)} \left( \widetilde{V} \left(x_{h+1}^{(\tau)} \right) -\left( \mathbb{P}_h \widetilde{V} \right) \left( x_h^{(\tau)}, a_h^{(\tau)} \right) \right)}_{(\star_1)} \\
        &\quad+\underbrace{\phi^\top\widetilde{\Lambda}^{-1}\zeta_{u,h}^{(t_m)}}_{(\star_2)}-\underbrace{\phi^\top\widetilde{\Lambda}^{-1}\bH z}_{(\star_3)}.
    \end{align*}

    We next bound the three terms separately. For $(\star_2)$, on $\cE_{\mathsf{noise}}$, we have $\norm{E_h^{(t_m)}}_{\mathsf{op}}\le\lambda/2$. Since $\Lambda_h^{(t_m)}\succeq 0$, it follows that $\widetilde{\Lambda}=\Lambda_h^{(t_m)}+\lambda I_d+E_h^{(t_m)}\succeq(\lambda/2)I_d$, and hence $\widetilde{\Lambda}^{-1}\preceq(2/\lambda)I_d$. Applying Cauchy--Schwarz in the inner product induced by $\widetilde{\Lambda}^{-1}$ and using $\norm{\zeta_{u,h}^{(t_m)}}_2\le\Sigma_u$, we obtain
    \begin{align*}
        \abs{(\star_2)} &\le \sqrt{ \phi^\top \widetilde{\Lambda}^{-1} \phi} \norm{ \zeta_{u,h}^{(t_m)}}_{\widetilde{\Lambda}^{-1}} \\
        &\le\Sigma_u\sqrt{\frac{2}{\lambda}}\sqrt{\phi^\top\widetilde{\Lambda}^{-1}\phi}.
    \end{align*}

    For $(\star_3)$, recall that $\bH=\lambda I_d+E_h^{(t_m)}$, so that $(\lambda/2) I_d \preceq \bH \preceq (3\lambda/2) I_d$ on $\cE_{\mathsf{noise}}$. Moreover, $\widetilde{\Lambda} = \Lambda_h^{(t_m)} + \bH \succeq \bH \succ 0$, which implies $\widetilde{\Lambda}^{-1} \preceq \bH^{-1}$. Consequently,
    \begin{align*}
        \norm{\bH z}_{\widetilde{\Lambda}^{-1}}^2
        &=z^\top\bH\widetilde{\Lambda}^{-1}\bH z
        \le z^\top\bH z
        \le\frac{3\lambda}{2}\norm{z}_2^2.
    \end{align*}
    Applying Cauchy--Schwarz and using $\norm{z}_2\le H\sqrt{d}$, we obtain
    \begin{align*}
        \abs{(\star_3)}
        &\le\sqrt{\phi^\top\widetilde{\Lambda}^{-1}\phi}\,\norm{\bH z}_{\widetilde{\Lambda}^{-1}} \\
        &\le H\sqrt{\frac{3\lambda d}{2}}\sqrt{\phi^\top\widetilde{\Lambda}^{-1}\phi}.
    \end{align*}

    We then consider $(\star_1)$. Here we adapt the proof of Lemma~B.3 in Appendix~C of~\citet{zhong2023theoreticalanalysisoptimisticproximal}. On $\cE_{\mathsf{noise}}$, we have $\lambda_{\min}\left(\widetilde{\Lambda}\right)\ge\frac{\lambda}{2}$ and 
    \begin{align*}
        \norm{\widetilde{w}_h^{(t_m)}}_2 &\le \norm{\widetilde{\Lambda}^{-1}}_{\mathsf{op}} \left( \sum_{\tau=1}^{t_m-1} \norm{\phi_h^{(\tau)}}_2 \abs{\widetilde{V}_{h+1}^{(t_m)}(x_{h+1}^{(\tau)})} + \norm{\zeta_{u,h}^{(t_m)}}_2 \right)\\ 
        &\le\frac{2}{\lambda}\left((t_m-1)H+\Sigma_u\right)\\ 
        &\le\frac{2(KH+\Sigma_u)}{\lambda}.
    \end{align*}

    In the proof, we first treat the sampled reward noise as fixed. We construct the value, $Q$-function, and policy classes following (C.2)--(C.4) of \citet{zhong2023theoreticalanalysisoptimisticproximal} and replace their batch rewards with $\widetilde{r}$ and their bonus coefficient $\beta$ with $\beta_{\mathsf{priv}}$. We allow each batch's average reward coefficient to range over the Euclidean ball of radius $L$. For the weights and matrices in these classes, we use the norm and eigenvalue bounds established above.

    On $\cE_{\mathsf{noise}}$, the estimated value function $\widetilde{V}_{h+1}^{(t_m)}$ belongs to the resulting value function class. Since $\norm{\psi(x,a)}_2 \le 1$ and clipping is non-expansive, the linear reward coefficients contribute an additional $pM$ parameters to the covering argument. We apply their Lemmas~C.1, C.2, and D.4 with covering accuracy $1/K$ for the value functions and covering accuracy $\left(2K+16\alpha MH^2K^2\right)^{-1}$ for the $Q$-functions. The log-covering number of the value function class at accuracy $1/K$ is then bounded by
    \begin{align*}
        \cO\left((d^2+p)M \iota \right),
    \end{align*}
    where $\iota=\log\left(\frac{c_1(d^2+p)HK(1+\alpha)(1+L)(1+\Sigma_u)}{\gamma}\right)$. To obtain this, we substitute the weight bound $2(KH+\Sigma_u)/\lambda$ and our choice of $\beta_{\mathsf{priv}}$ into the covering bounds and then use $M \le K$ and $\lambda \ge 1$ to simplify the logarithmic terms.

    We now apply their Lemma~D.3 to the value function class, using the regularized matrix $\Lambda_h^{(t_m)}+\lambda I_d$ and covering accuracy $1/K$. For each pair $(m,h)$, we set the failure probability to $\gamma/(MH)$. For each fixed realization of the reward noise, a union bound over all $MH$ pairs gives a total failure probability of at most $\gamma$.

    After integrating over the reward perturbations and intersecting with $\cE_{\mathsf{noise}}$, with probability at least $1-2\gamma$,
    \begin{align*}
        \norm{ \sum_{\tau=1}^{t_m-1} \phi_h^{(\tau)} \left(\widetilde{V}\left(x_{h+1}^{(\tau)}\right)-\left(\mathbb{P}_h \widetilde{V}\right) \left( x_h^{(\tau)}, a_h^{(\tau)}\right) \right)}_{\left(\Lambda_h^{(t_m)}+\lambda I_d\right)^{-1}} 
        \le c_2H\sqrt{(d^2+p)M\iota},
    \end{align*}
    simultaneously for all $(m,h)$, where $c_2>0$ is an absolute constant.

    On $\cE_{\mathsf{noise}}$, we have $\widetilde{\Lambda}\succeq\frac12\left(\Lambda_h^{(t_m)}+\lambda I_d\right)$, and hence $\widetilde{\Lambda}^{-1}\preceq2\left(\Lambda_h^{(t_m)}+\lambda I_d\right)^{-1}$.
    Thus, the norm induced by $\widetilde{\Lambda}^{-1}$ is at most $\sqrt{2}$ times the norm induced by $\left(\Lambda_h^{(t_m)}+\lambda I_d\right)^{-1}$. Applying Cauchy--Schwarz and choosing $c_1 \ge\sqrt{2} c_2$, we obtain
    \begin{align*}
        \abs{(\star_1)}
        &\le\sqrt{2}c_2H\sqrt{(d^2+p)M\iota}
        \sqrt{\phi^\top\widetilde{\Lambda}^{-1}\phi}.
    \end{align*}
    
    Combining the bounds for $(\star_1)$, $(\star_2)$, and $(\star_3)$ proves the lemma.
    
\end{proof}

We then show that the estimate remains optimistic after clipping and that its error is bounded by the bonus.
\begin{lemma} \label{lem:private-transition-optimism}
    With $\beta_{\mathsf{priv}}$ chosen as in Lemma~\ref{lem:private-confidence}, the following holds with probability at least $1-2\gamma$, simultaneously for all $m\in[M]$, $h\in[H]$, and $(x,a)\in\cS\times\cA$:
    \begin{align*}
        -2\min\left\{H,\widetilde{\Gamma}_h^{(t_m)}(x,a)\right\}&\le\left(\mathbb{P}_h\widetilde{V}_{h+1}^{(t_m)}-\widehat{\mathbb{P}}_h^{(t_m)}\widetilde{V}_{h+1}^{(t_m)}\right)(x,a)\le0.
    \end{align*}
\end{lemma}

\begin{proof}
    On the event in Lemma~\ref{lem:private-confidence}, the definition of the bonus $\widetilde{\Gamma}_h^{(t_m)}$ gives
    \begin{align*}
        \left( \mathbb{P}_h \widetilde{V}_{h+1}^{(t_m)} \right)(x,a)
        &\le \phi(x,a)^\top\widetilde{w}_h^{(t_m)}+\widetilde{\Gamma}_h^{(t_m)}(x,a) \\
        &\le \left(\mathbb{P}_h\widetilde{V}_{h+1}^{(t_m)}\right)(x,a)+2\widetilde{\Gamma}_h^{(t_m)}(x,a).
    \end{align*}

    Recall that we have $\widetilde{V}_{h+1}^{(t_m)} \in [0,H-h]$, its expectation $\left( \mathbb{P}_h \widetilde{V}_{h+1}^{(t_m)} \right)(x,a)$ is also in this interval. The prediction $\phi(x,a)^\top \widetilde{w}_h^{(t_m)} + \widetilde{\Gamma}_h^{(t_m)}(x,a)$ is already at least this expectation, so it is nonnegative. If the prediction exceeds $H-h$, clipping lowers it to $H-h$, which is still at least the expectation. Otherwise, clipping leaves it unchanged. Hence,
    \begin{align*}
        0 \le \left(\widehat{\mathbb{P}}_h^{(t_m)} \widetilde{V}_{h+1}^{(t_m)} -\mathbb{P}_h \widetilde{V}_{h+1}^{(t_m)} \right)(x,a) \le 2 \widetilde{\Gamma}_h^{(t_m)}(x,a).
    \end{align*}

    The clipped prediction and the true expected value both lie in $[0,H-h]$, so the prediction can exceed the expectation by at most $H$. We have already shown that this gap is at most $2 \widetilde{\Gamma}_h^{(t_m)}(x,a)$. Thus, the gap is at most $\min\left\{H,2\widetilde{\Gamma}_h^{(t_m)}(x,a)\right\}$, which is no larger than $2\min\left\{H,\widetilde{\Gamma}_h^{(t_m)}(x,a)\right\}$. Finally, changing the sign gives the desired bound.
    
\end{proof}

The optimism bound shows that the transition term under $\pi^\star$ is nonpositive. It also bounds each transition error along the learner's trajectory by twice the bonus clipped at $H$. Substituting these bounds into the Bellman-error bound in~\eqref{eq:private-bellman} and applying a union bound, we obtain, with probability at least $1-4\gamma$,
\begin{align}
    &\sum_{k=1}^{K} \sum_{h=1}^{H} \left( \mathbb{E}_{\pi^\star} \left[ \widetilde{\delta}_h^{(k)}(x_h,a_h)\right] - \mathbb{E}_{\pi^{(k)}} \left[ \widetilde{\delta}_h^{(k)}(x_h,a_h)\right] \right) \notag \\
    &\quad \le 2\sum_{m=1}^{M} \sum_{k=t_m}^{t_{m+1}-1} \sum_{h=1}^{H} \min\left\{H,\widetilde{\Gamma}_h^{(t_m)} \left(x_h^{(k)},a_h^{(k)}\right) \right\} \notag \\
    &\qquad + BH + \alpha H^3K + 3\sqrt{2H^3K\log\left(\frac{2}{\gamma} \right)} + 2KH\bar{\zeta}_r. \label{eq:private-bellman-bonus-bound}
\end{align}

It remains to bound the sum of clipped bonuses. We adapt the proof of Lemma~A.2 and Claim~A.6 of \citet{chowdhury2022shuffle}. We first introduce a virtual design matrix that uses all data available before each episode and contains no noise and then compare the bonuses computed at batch starts with those computed before each episode.

\begin{lemma} \label{lem:private-bonus-sum} Take $\lambda\ge\max\{1,2\Sigma_E\}$.
    On $\cE_{\mathsf{noise}}$, we have
    \begin{align*}
        &\sum_{m=1}^{M} \sum_{k=t_m}^{t_{m+1}-1} \sum_{h=1}^{H} \min\left\{H, \widetilde{\Gamma}_h^{(t_m)} \left(x_h^{(k)},a_h^{(k)}\right)\right\} \\
        &\quad \le \frac{H^2Bd}{2\log 2} \log \left(1+\frac{2K}{d\lambda}\right) \\
        &\qquad + 2H \beta_{\mathsf{priv}} \sqrt{2dK\log\left(1+\frac{2K}{d\lambda}\right)}.
    \end{align*}
\end{lemma}

\begin{proof}
    On $\cE_{\mathsf{noise}}$, for each $h\in[H]$ and $k\in[K+1]$, define the virtual design matrix
    \begin{align*}
        \bar{\Lambda}_h^{(k)}:= \sum_{\tau=1}^{k-1} \phi_h^{(\tau)} \left(\phi_h^{(\tau)} \right)^\top + \frac{\lambda}{2}I_d.
    \end{align*}

    This matrix uses the data from all episodes before $k$, without adding noise. Let
    \begin{align*}
        \bar{\Gamma}_h^{(k)}(x,a) := \beta_{\mathsf{priv}} \sqrt{\phi(x,a)^\top \left( \bar{\Lambda}_h^{(k)} \right) ^{-1} \phi(x,a)}
    \end{align*}
    be the bonus computed from this matrix. Since $\norm{E_h^{(t_m)}}_{\mathsf{op}}\le\lambda/2$, we have
    \begin{align*}
        \widetilde{\Lambda}_h^{(t_m)} \succeq \bar{\Lambda}_h^{(t_m)}, \quad \widetilde{\Gamma}_h^{(t_m)}(x,a) \le \bar{\Gamma}_h^{(t_m)} (x,a).
    \end{align*}

    Following the proof of Claim~A.6 of~\citet{chowdhury2022shuffle}, we consider the batches in which the log-determinant increases by more than $2\log 2$. Define
    \begin{align*}
        \cB_h := \left\{m\in[M]: \frac{\det\left(\bar{\Lambda}_h^{(t_{m+1})}\right)}{\det \left( \bar{\Lambda}_h^{(t_m)} \right)}>4 \right\}.
    \end{align*}

    Each new data point adds a positive semidefinite matrix to the virtual design matrix, so its determinant cannot decrease. Summing the log-determinant increases over all batches, we obtain
    \begin{align*}
        2\log 2 \cdot \abs{\cB_h}&\le\sum_{m=1}^{M} \log \left( \frac{ \det \left( \bar{\Lambda}_h^{(t_{m+1})} \right)}{\det \left(\bar{\Lambda}_h^{(t_m)}\right)}\right) \\
        & = \log \left( \frac{ \det \left( \bar{\Lambda}_h^{(K+1)}\right)}{\det \left( \bar{\Lambda}_h^{(1)} \right)} \right) \\
        &\le d\log \left(1+\frac{2K}{d\lambda} \right),
    \end{align*}
    where the last inequality comes from the trace-determinant lemma (cf. Lemma 10 in~\citet{abbasi2011improved}).

    Each batch in $\cB_h$ contains $B$ episodes, and each clipped bonus is at most $H$. Thus, the total contribution of these batches is at most $BH\abs{\cB_h}$.

    For a batch $m \notin \cB_h$ and any episode $t_m \le k <t_{m+1}$, we have $\bar{\Lambda}_h^{(t_m)} \preceq \bar{\Lambda}_h^{(k)}$ and $\det\left( \bar{\Lambda}_h^{(k)} \right) / \det \left( \bar{\Lambda}_h^{(t_m)} \right) \le 4$. Applying the matrix norm inequality used in the proof of Claim~A.6 of \citet{chowdhury2022shuffle} to the inverse matrices gives
    \begin{align*}
        \bar{\Gamma}_h^{(t_m)}(x,a) \le \sqrt{ \frac{\det\left(\bar{\Lambda}_h^{(k)} \right)}{\det \left( \bar{\Lambda}_h^{(t_m)} \right)}} \bar{\Gamma}_h^{(k)}(x,a) \le 2 \bar{\Gamma}_h^{(k)}(x,a).
    \end{align*}
    
    Thus, in these batches, the bonus $\bar{\Gamma}_h^{(t_m)}$ computed at the batch start is at most twice $\bar{\Gamma}_h^{(k)}$, which uses all data available before episode $k$.

    We next bound the sum of $\bar{\Gamma}_h^{(k)}\left(x_h^{(k)},a_h^{(k)}\right)$ over all episodes, using the elliptical potential argument as in the proof of Lemma~A.2 of \citet{chowdhury2022shuffle}. Since $\bar{\Lambda}_h^{(k)} \succeq (\lambda/2) I_d$, $\norm{\phi_h^{(k)}}_2 \le 1$, and $\lambda \ge 1$, we have
    \begin{align*}
        \norm{\phi_h^{(k)}}_{\left(\bar{\Lambda}_h^{(k)}\right)^{-1}}^2 \le \frac{2}{\lambda} \le 2.
    \end{align*}
    
    Using $u \le 2\log\left(1+u\right)$ for $u\in[0,2]$ and the matrix determinant lemma, we obtain
    \begin{align*}
        \sum_{k=1}^{K} \norm{ \phi_h^{(k)} }_{\left( \bar{\Lambda}_h^{(k)} \right)^{-1}}^2 
        &\le 2 \sum_{k=1}^{K} \log \left( 1 + \norm{\phi_h^{(k)}}_{\left(\bar{\Lambda}_h^{(k)} \right)^{-1}}^2 \right) \\
        & = 2 \log \left( \frac{\det\left(\bar{\Lambda}_h^{(K+1)}\right)}{\det \left( \bar{\Lambda}_h^{(1)} \right)} \right) \\
        &\le2d\log\left(1+\frac{2K}{d\lambda}\right).
    \end{align*}
    
    By Cauchy--Schwarz,
    \begin{align*}
        \sum_{k=1}^{K} \bar{\Gamma}_h^{(k)} \left(x_h^{(k)},a_h^{(k)}\right) &=\beta_{\mathsf{priv}}\sum_{k=1}^{K}\norm{\phi_h^{(k)}}_{\left(\bar{\Lambda}_h^{(k)}\right)^{-1}} \\
        &\le \beta_{\mathsf{priv}} \sqrt{K\sum_{k=1}^{K} \norm{ \phi_h^{(k)} }_{ \left(\bar{\Lambda}_h^{(k)} \right)^{-1}}^2} \\
        &\le \beta_{\mathsf{priv}} \sqrt{2dK \log \left(1+\frac{2K}{d\lambda}\right)}.
    \end{align*}
    
    Combining the bounds for batches in $\cB_h$ and not in $\cB_h$, together with $\widetilde{\Gamma}_h^{(t_m)} \le \bar{\Gamma}_h^{(t_m)}$, yields
    \begin{align*}
        &\sum_{m=1}^{M} \sum_{k=t_m}^{t_{m+1}-1} \min \left\{H,\widetilde{\Gamma}_h^{(t_m)} \left(x_h^{(k)}, a_h^{(k)} \right)\right\} \\
        &\quad\le BH \abs{\cB_h} + 2\sum_{k=1}^{K}\bar{\Gamma}_h^{(k)} \left(x_h^{(k)},a_h^{(k)}\right) \\
        &\quad \le \frac{HBd}{2\log 2} \log \left(1+\frac{2K}{d\lambda}\right) + 2\beta_{\mathsf{priv}} \sqrt{2dK\log\left(1+\frac{2K}{d\lambda}\right)}.
    \end{align*}
    Summing over $h\in[H]$ proves the lemma.
    
\end{proof}

\begin{lemma}[Privacy guarantee]\label{lem:privacy-calibration}
Fix $\epsilon\in(0,1]$ and $\delta\in(0,1)$. Set
\begin{align*}
    \sigma_u &:=\frac{2H \sqrt{MH}\sqrt{2\log\left(3.75/\delta\right)}}{\epsilon/3}=\widetilde{\cO}\left(\frac{H^{3/2}\sqrt{M}}{\epsilon}\right), \\
    \sigma_{\mathsf{node}}&:=\frac{2\sqrt{H\left(1+ \log_2 K\right)}\sqrt{2\log\left(3.75/\delta\right)}}{\epsilon/3}, \\
    \sigma_\Lambda&=\sigma_{\mathsf{node}}\sqrt{1+\log_2 K}=\widetilde{\cO}\left(\frac{\sqrt{H}\log(2K)}{\epsilon}\right), \\
    \sigma_\theta&=\frac{2L\sqrt{H}\sqrt{2\log\left(3.75/\delta\right)}}{B\left(\epsilon/3\right)}=\widetilde{\cO}\left(\frac{L\sqrt{H}}{B\epsilon}\right).
\end{align*}
Then Algorithm~\ref{alg:privPO} is $(\epsilon,\delta)$-JDP under the neighboring relation in the preliminaries. Here $\sigma_\Lambda$ bounds the standard deviation of each entry of $E_h^{(t_m)}$.
\end{lemma}

\begin{proof}
    Following the proof of Theorem~6.4 in~\citet{zhou2022differentially}, we analyze the Gaussian releases using the zero-concentrated differential privacy (zCDP)~\citep{bun2016concentrateddifferentialprivacysimplifications} and then convert the result to $(\epsilon, \delta)$-DP. We split the privacy budget as $(\epsilon/3,\delta/3)$ among $\{\widetilde{u}\}$, $\{\widetilde{\Lambda}\}$, and $\{\widetilde{r}\}$ in the following proof.

    Fix two user sequences differing only in user $i$, and fix the random seeds used for the adversary and action sampling. Under the same preceding private outputs, the policies used in completed episodes are fixed. Each unchanged user then receives the same actions and gives the same state responses and reward choices in the two interactions, so only episode $i$ can contribute different data. All sensitivity bounds below hold for every preceding output history and every fixed choice of the action-sampling seeds.

    We start with the matrix releases $\widetilde{\Lambda}$. Each episode contributes a matrix $\phi\phi^\top$ to a tree node, and replacing that contribution changes the node sum by at most
    \begin{align*}
        \norm{\phi\phi^\top-\phi'\left(\phi'\right)^\top}_{\mathsf{F}}\le\norm{\phi}_2^2+\norm{\phi'}_2^2\le2.
    \end{align*}
    The same bound holds for the Euclidean norm of the upper-triangular entries. Adding independent Gaussian noise of variance $\sigma_{\mathsf{node}}^2$ to these entries therefore gives $\left(2/\sigma_{\mathsf{node}}^2\right)$-zCDP for each affected node. Only nodes whose intervals contain episode $i$ can differ. Since there are at most $1+ \log_2K$ such nodes per stage, their total contribution to the zCDP parameter is at most
    \begin{align*}
        \rho_\Lambda := \frac{2H\left(1+\log_2K\right)}{\sigma_{\mathsf{node}}^2}=\frac{\epsilon^2}{36\log\left(3.75/\delta\right)}.
    \end{align*}
    Each prefix matrix sums at most $1+\log_2K$ noisy nodes; their independent perturbations give the stated bound $\sigma_\Lambda$. Each noisy tree node is counted once in the privacy composition; reusing it to form prefix matrices is post-processing.

    We next consider the target vectors $\widetilde{u}$. The preceding private outputs fix $\widetilde{V}_{h+1}^{(t_m)}$, so replacing user $i$ changes at most one summand of the unperturbed target $u_h^{(t_m)}:=\widetilde{u}_h^{(t_m)}-\zeta_{u,h}^{(t_m)}$. Since $\norm{\phi(x,a)}_2\le1$ and $0\le\widetilde{V}_{h+1}^{(t_m)}\le H$, the sensitivity is at most $2H$. These vectors are recomputed using the current value function at each batch start, yielding at most $MH$ Gaussian releases and a total zCDP contribution of
    \begin{align*}
        \rho_u:=\frac{MH(2H)^2}{2\sigma_u^2}\le\frac{\epsilon^2}{36\log\left(3.75/\delta\right)}.
    \end{align*}
    The value-function bound follows from clipping and holds for every realization of the privacy perturbations.

    For the private rewards $\widetilde{r}$, we add Gaussian noise to the batch-average coefficient vectors. Since $\norm{\theta_h^{(k)}}_2 \le L$, replacing one user's coefficient vector at a fixed stage changes it by at most $2L$ in Euclidean norm. After averaging over the batch, the sensitivity is at most $2L/B$. A user's rewards enter only one batch average, so at most $H$ stage-wise releases can differ. Their total zCDP contribution is at most
    \begin{align*}
        \rho_r:=\frac{H(2L/B)^2}{2\sigma_\theta^2}=\frac{\epsilon^2}{36\log\left(3.75/\delta\right)}.
    \end{align*}
    Reconstructing the rewards from the noisy coefficients and clipping them are post-processing.

    We now add up the privacy costs of the matrix, target-vector, and reward releases. For each release, we compare the two neighboring inputs while keeping all previous outputs the same. The Gaussian zCDP bound holds for every such history, so we can combine these bounds in the actual release order and account for each type separately. Applying the tail bound in the proof of Lemma~3.5 of~\citet{bun2016concentrateddifferentialprivacysimplifications}, each type $j\in\left\{\Lambda,u,r\right\}$ contributes at most
    \begin{align*}
        \rho_j+2\sqrt{\rho_j\log\left(3/\delta\right)}\le\frac{\epsilon}{3}
    \end{align*}
    to the total privacy loss with probability at least $1-\delta/3$. The inequality follows from the bounds on $\rho_j$ and $\epsilon\le1$. A union bound shows that all three bounds hold together with probability at least $1-\delta$. On this event, their contributions add up to at most $\epsilon$. Thus, the joint transcript is $(\epsilon,\delta)$-DP.

    Finally, we apply the billboard lemma~\citep{hsu2016}. The DP bound above holds for every fixed choice of the random seeds used to sample actions. Since these seeds are independent of the users' private inputs, the same DP guarantee holds when the seeds are released together with the private statistics. Given these statistics and seeds, the actions recommended to each user depend only on that user's own input. Selecting the final policy $\widehat{\pi}$ also uses only the private statistics and independent randomness. The billboard lemma and post-processing therefore imply that Algorithm~\ref{alg:privPO} is $(\epsilon,\delta)$-JDP.

\end{proof}

We now use these noise scales to specify the three bounds left open in the regret analysis: $\Sigma_E$, $\Sigma_u$, and $\bar{\zeta}_r$.

\begin{lemma} \label{lem:private-noise-bounds}
    Under the noise generation and scales in Lemma~\ref{lem:privacy-calibration}, fix $\gamma\in(0,1)$ and set
    \begin{align*}
        \Sigma_E&:= c_3 \cdot \sigma_\Lambda \left(\sqrt{d}+\sqrt{\log\left(\frac{4MH}{\gamma}\right)}\right), \\
        \Sigma_u&:=\sigma_u\left(\sqrt{d}+\sqrt{2\log\left(\frac{2MH}{\gamma}\right)}\right), \\
        \bar{\zeta}_r&:=\sigma_\theta \sqrt{2\log\left(\frac{2MHSA}{\gamma}\right)},
    \end{align*}
    where $c_3>0$ is a sufficiently large absolute constant. With probability at least $1-\gamma$,
    \begin{align*}
        \norm{E_h^{(t_m)}}_{\mathsf{op}}\le\Sigma_E, \qquad \norm{\zeta_{u,h}^{(t_m)}}_2\le\Sigma_u
    \end{align*}
    simultaneously for all $m\in[M]$ and $h\in[H]$; that is, $\cE_{\mathsf{noise}}$ holds. Separately, with probability at least $1-\gamma$,
    \begin{align*}
        \norm{\widetilde{r}_h^{(t_m)}-\bar{r}_h^{(t_m)}}_\infty\le\bar{\zeta}_r
    \end{align*}
    simultaneously for all $m\in[M]$ and $h\in[H]$.
\end{lemma}

\begin{proof}
    We first consider the matrix perturbations. For each fixed $(m,h)$, the entries of $E_h^{(t_m)}$ on and above the diagonal are independent centered Gaussian random variables with variance at most $\sigma_\Lambda^2$, and the matrix is symmetric. Thus, we can use the same symmetric-matrix concentration bound as in the proof of Lemma~A.4 of~\citet{chowdhury2022shuffle}, which gives
    \begin{align*}
        \mathbb{P}\left(\norm{E_h^{(t_m)}}_{\mathsf{op}} > c_3 \sigma_\Lambda \left(\sqrt{d} + \sqrt{\log\left(\frac{4MH}{\gamma}\right)}\right)\right)\le\frac{\gamma}{2MH}.
    \end{align*}
    
    A union bound over the $MH$ batch-stage pairs shows that all matrix bounds hold with probability at least $1-\gamma/2$. The matrices at different batch starts may share tree noise, but the union bound does not require them to be independent.
    
    For the target-vector noise, $\zeta_{u,h}^{(t_m)}\sim \cN \left(0,\sigma_u^2I_d\right)$. The Gaussian norm bound gives
    \begin{align*}
        \mathbb{P} \left( \norm{\zeta_{u,h}^{(t_m)}}_2 >\sigma_u \left(\sqrt{d} + \sqrt{2\log\left(\frac{2MH}{\gamma}\right)}\right)\right) 
        \le \frac{\gamma}{2MH}.
    \end{align*}
    Taking a union bound over all $(m,h)$ gives the vector bounds with probability at least $1-\gamma/2$. Combining the matrix and vector bounds proves $\Pr\left(\cE_{\mathsf{noise}}\right)\ge 1-\gamma$.

    Finally, the reward perturbations are feature projections of the coefficient noise. For every fixed $(m,h,x,a)$,
    \begin{align*}
        \zeta_{r,h}^{(t_m)}(x,a)=\inner{\psi(x,a)}{\zeta_{\theta,h}^{(t_m)}}\sim\cN\left(0,\sigma_{\theta}^2\norm{\psi(x,a)}_2^2\right).
    \end{align*}
    The variance is at most $\sigma_{\theta}^2$. The two-sided Gaussian tail bound therefore gives
    \begin{align*}
        \mathbb{P} \left(\abs{\zeta_{r,h}^{(t_m)}(x,a)}>\bar{\zeta}_r\right)\le2\exp\left(-\frac{\bar{\zeta}_r^2}{2\sigma_{\theta}^2}\right)=\frac{\gamma}{MHSA}.
    \end{align*}
    A union bound over all $MHSA$ batch-stage-state-action tuples gives $\norm{\zeta_{r,h}^{(t_m)}}_\infty\le\bar{\zeta}_r$ for every $(m,h)$ with probability at least $1-\gamma$. The perturbations at different state-action pairs can be correlated; independence is not needed for this union bound. On an infinite state space the same reward error is bounded by $\|\zeta_{\theta,h}^{(t_m)}\|_2$ since $\|\psi(x,a)\|_2\le1$. This controls the reward perturbation uniformly with a Gaussian norm bound depending on $p$ and the failure probability. Since $\bar{r}_h^{(t_m)}(x,a)\in[0,1]$, clipping cannot increase its distance from the batch average. Hence,
    \begin{align*}
        \norm{\widetilde{r}_h^{(t_m)}-\bar{r}_h^{(t_m)}}_\infty\le\norm{\zeta_{r,h}^{(t_m)}}_\infty\le\bar{\zeta}_r,
    \end{align*}
    which proves the reward bound.

\end{proof}

Substituting the scales from Lemma~\ref{lem:privacy-calibration} yields
\begin{align*}
    \Sigma_E =\widetilde{\cO}\left(\frac{\sqrt{dH}}{\epsilon}\right), \quad
    \Sigma_u =\widetilde{\cO}\left(\frac{H^{3/2}\sqrt{dM}}{\epsilon}\right), \quad
    \bar{\zeta}_r =\widetilde{\cO}\left(\frac{L\sqrt{H}}{B\epsilon}\right).
\end{align*}
Here $\widetilde{\cO}$ hides logarithmic factors in $K,H,S,A,1/\delta$, and $1/\gamma$. In particular, the reward-noise contribution in \eqref{eq:private-bellman} satisfies
\begin{align*}
    2KH\bar{\zeta}_r=\widetilde{\cO}\left(\frac{H^{3/2}LK}{B\epsilon}\right).
\end{align*}
Choosing $\lambda\ge\max\left\{1,2\Sigma_E\right\}$ now ensures $\norm{E_h^{(t_m)}}_{\mathsf{op}}\le\lambda/2$ on $\cE_{\mathsf{noise}}$, as required by the confidence and bonus bounds.

\begin{proof}[Remaining Proof of Theorem~\ref{thm:reg}]

We apply the previous high probability bounds with $\gamma / 4$ in place of $\gamma$. In particular, $\Sigma_E$, $\Sigma_u$, $\bar{\zeta}_r$, and $\beta_{\mathsf{priv}}$ are chosen as in Lemmas~\ref{lem:private-noise-bounds} and~\ref{lem:private-confidence} with the same substitution. The Bellman-error bound in~\eqref{eq:private-bellman-bonus-bound} and the bonus bound in Lemma~\ref{lem:private-bonus-sum} then hold with probability at least $1-\gamma$.

Now we combine these terms. Substituting Lemma~\ref{lem:private-policy-error}, \eqref{eq:private-bellman-bonus-bound}, and Lemma~\ref{lem:private-bonus-sum} into~\eqref{eq:private-regret-decomposition} gives
\begin{align*}
    \reg(K)&\le\frac{\alpha H^3K}{2}+\frac{BH\log A}{\alpha}+BH+\alpha H^3K \\
    &\quad+ 3\sqrt{2H^3K\log\left(\frac{8}{\gamma}\right)}+2KH\bar{\zeta}_r \\
    &\quad+2\left(\frac{H^2Bd}{2\log 2}\log\left(1+\frac{2K}{d\lambda}\right)+2H\beta_{\mathsf{priv}}\sqrt{2dK\log\left(1+\frac{2K}{d\lambda}\right)}\right),
\end{align*}
where $\beta_{\mathsf{priv}}$ is given in Lemma~\ref{lem:private-confidence}, and Lemma~\ref{lem:private-noise-bounds} gives
\begin{align*}
    \Sigma_u = \widetilde{\cO}\left(\frac{H^{3/2}\sqrt{dM}}{\epsilon}\right), \quad 2KH\bar{\zeta}_r = \widetilde{\cO}\left(\frac{H^{3/2}LK}{B\epsilon}\right).
\end{align*}

The confidence radius also contributes logarithms in $d,p,1+L$ and $1/\epsilon$ through $\iota$. Using the choice of $\alpha$ below and ignoring log factors, we obtain
\begin{align}
    \reg(K)&\le\widetilde{\cO}\left(H^2\sqrt{BK}+H^2dB+\frac{H^2\sqrt{d(d^2+p)}K}{\sqrt{B}}\right. \notag \\
    &\qquad\left.+H^2d\sqrt{\lambda K}+\frac{H^{5/2}dK}{\epsilon\sqrt{B\lambda}}+\frac{H^{3/2}L K}{B\epsilon}\right). \label{eq:private-regret-before-tuning}
\end{align}

Balancing the two $\lambda$-dependent terms up to absolute constants gives
\begin{align*}
    \lambda^\star&:=\frac{\Sigma_u}{H\sqrt{d}}=\widetilde{\cO}\left(\frac{\sqrt{H}}{\epsilon}\sqrt{\frac{K}{B}}\right).
\end{align*}
To ensure $\norm{E_h^{(t_m)}}_{\mathsf{op}}\le\lambda/2$ on $\cE_{\mathsf{noise}}$ and $\lambda\ge1$, we choose
\begin{align*}
    \lambda :=\max\left\{1,2\Sigma_E,\lambda^\star\right\}, \quad \Sigma_E = \widetilde{\cO}\left(\frac{\sqrt{dH}}{\epsilon}\right).
\end{align*}
Therefore,
\begin{align*}
    H^2d\sqrt{\lambda K}+\frac{H^{5/2}dK}{\epsilon\sqrt{B\lambda}} \le \widetilde{\cO} \left(\frac{H^{9/4}dK^{3/4}}{\sqrt{\epsilon}B^{1/4}}+\frac{H^{9/4}d^{5/4}K^{1/2}}{\sqrt{\epsilon}}\right).
\end{align*}
Here $\lambda\ge\lambda^\star$ controls the term containing $1/\sqrt{\lambda}$. The contribution of the lower bound $\lambda\ge1$ is absorbed by the second term on the right since $\epsilon\le1$ and $H,d\ge1$.

We choose the following parameters
\begin{align*}
    \alpha&:=\frac{1}{H}\sqrt{\frac{2B\log A}{3K}}, \quad B:=\sqrt{d^3K}.
\end{align*}

This choice of batch size satisfies $B\le K$ only when $K\ge d^3$. If $K<d^3$, then $\sqrt{d^3K}>K$, so we instead take $B=K$ and use a single batch. At $k=1$, the initialization $\widetilde{Q}_h^{(0)}\equiv0$ leaves the uniform policy unchanged in the policy-improvement step, and no further policy updates occur. Privacy of the public releases follows from Lemma~\ref{lem:privacy-calibration} in this case as well. Since each episode's total reward lies in $[0,H]$, we have deterministically
\begin{align*}
    \reg(K)\le HK\le H^2d^{3/4}K^{3/4}.
\end{align*}
This establishes the claimed regret and privacy guarantees when $K<d^3$. It remains to consider $K\ge d^3$, for which we use $B=\sqrt{d^3K}$ as derived above.

Substituting these choices, we finally obtain
\begin{align*}
    \reg(K) &\le \widetilde{\cO} \left(H^2 d^{3/4}\sqrt{1+\frac{p}{d^2}} K^{3/4}+H^2d^{5/2}K^{1/2} + \frac{H^{9/4}d^{5/8}K^{5/8}}{\sqrt{\epsilon}} \right. \\
    &\left.\quad+ \frac{H^{9/4}d^{5/4}K^{1/2}}{\sqrt{\epsilon}}+\frac{H^{3/2}LK^{1/2}}{d^{3/2}\epsilon} \right).
\end{align*}

The regret bound holds with probability at least $1-\gamma$, and Lemma~\ref{lem:privacy-calibration} gives $(\epsilon,\delta)$-JDP. This completes the proof.

\end{proof}

\end{document}